\documentclass[nohyperref]{article}

\usepackage{microtype}
\usepackage{graphicx}
\usepackage{subfigure}
\usepackage{booktabs} % for professional tables

\usepackage{hyperref}
\hypersetup{
hidelinks,
colorlinks=true,
linkcolor=black,
citecolor=blue
}
\usepackage{multirow}
\usepackage{amsmath}
\usepackage{amssymb}
\usepackage{array}

\usepackage[accepted]{icml2023}

\usepackage{amsmath}
\usepackage{amssymb}
\usepackage{mathtools}
\usepackage{amsthm}

\usepackage[capitalize,noabbrev]{cleveref}

\theoremstyle{plain}

\theoremstyle{definition}

\theoremstyle{remark}

\usepackage[textsize=tiny]{todonotes}

\icmltitlerunning{Submission of MTS-Mixers for ICML 2023}

\begin{document}

\twocolumn[
\icmltitle{MTS-Mixers: Multivariate Time Series Forecasting \\ 
via Factorized Temporal and Channel Mixing}




\begin{icmlauthorlist}
\icmlauthor{Zhe Li}{loc:a}
\icmlauthor{Zhongwen Rao}{loc:b}
\icmlauthor{Lujia Pan}{loc:b}
\icmlauthor{Zenglin Xu}{loc:a}
\end{icmlauthorlist}

\icmlaffiliation{loc:a}{Harbin Institute of Technology, Shenzhen}
\icmlaffiliation{loc:b}{Huawei Technologies Ltd.}

\icmlcorrespondingauthor{Zhe Li}{plum271828@gmail.com}
\icmlcorrespondingauthor{Zhongwen Rao}{raozhongwen@huawei.com}
\icmlcorrespondingauthor{Lujia Pan}{panlujia@huawei.com}
\icmlcorrespondingauthor{Zenglin Xu}{zenglin@gmail.com}

\icmlkeywords{Machine Learning, ICML}

\vskip 0.3in
]



\printAffiliationsAndNotice{}  % leave blank if no need to mention equal contribution

\begin{abstract}
Multivariate time series forecasting has been widely used in various practical scenarios. Recently, Transformer-based models have shown significant potential in forecasting tasks due to the capture of long-range dependencies. However, recent studies in the vision and NLP fields show that the role of attention modules is not clear, which can be replaced by other token aggregation operations. This paper investigates the contributions and deficiencies of attention mechanisms on the performance of time series forecasting. Specifically, we find that (1) attention is not necessary for capturing temporal dependencies, (2) the entanglement and redundancy in the capture of temporal and channel interaction affect the forecasting performance, and (3) it is important to model the mapping between the input and the prediction sequence. To this end, we propose MTS-Mixers, which use two factorized modules to capture temporal and channel dependencies. Experimental results on several real-world datasets show that MTS-Mixers outperform existing Transformer-based models with higher efficiency.
\end{abstract}

\section{Introduction}\label{sec:1}
% Note use of \abovespace and \belowspace to get reasonable spacing
% above and below tabular lines.

\begin{figure}[ht]
% \vskip 0.2in
\begin{center}
\centerline{\includegraphics[width=0.8\columnwidth]{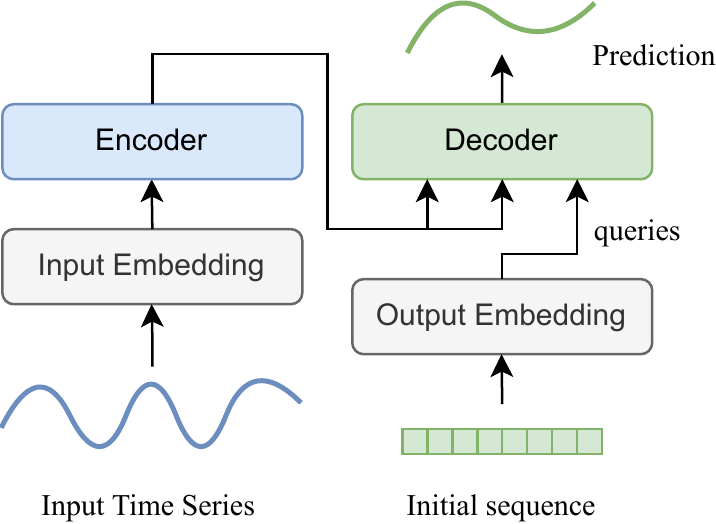}}
\caption{The overall architecture of Transformer-based models for time series forecasting. Notice that the generation of the prediction sequence in the decoder is non-autoregressive.}
\label{fig:1}
\end{center}
\vskip -0.2in
\end{figure}

Multivariate time series forecasting has been an increasingly popular topic in various scenarios, such as electricity forecasting~\cite{Khan2020TowardsEE}, weather forecasting~\cite{Angryk2020MultivariateTS}, and traffic flow estimation~\cite{Chen2001FreewayPM}. With the scaling of computing resources and model architectures, deep learning techniques, including RNN-based~\cite{Lai2018ModelingLA} and CNN-based models~\cite{Bai2018AnEE}, have achieved better prediction performance than traditional statistical methods~\cite{Zhang2003TimeSF,Ariyo2014StockPP}. Due to the capture of long-range dependencies, Transformer~\cite{Vaswani2017AttentionIA} has been recently used to catch long-term temporal correlation in time series forecasting and has demonstrated promising results~\cite{Zhou2021InformerBE, Wu2021AutoformerDT,Liu2022PyraformerLP,Zhou2022FEDformerFE}. The overall architecture of existing Transformer-based models on time series forecasting is illustrated in Figure~\ref{fig:1}. In the embedding module, the input time series are generally fed into one 1D convolutional layer to generate temporal tokens with positional or date-specific encoding at the point level for preserving ordering information. Then the encoder will capture the temporal interaction among time points via an attention-like mechanism, and the decoder will generate the prediction sequence in one pass.

\begin{figure*}[ht]
\vskip 0.2in
\centering
\subfigure[Modifications of Transformer.\label{fig:2a}]{\includegraphics[scale=0.4]{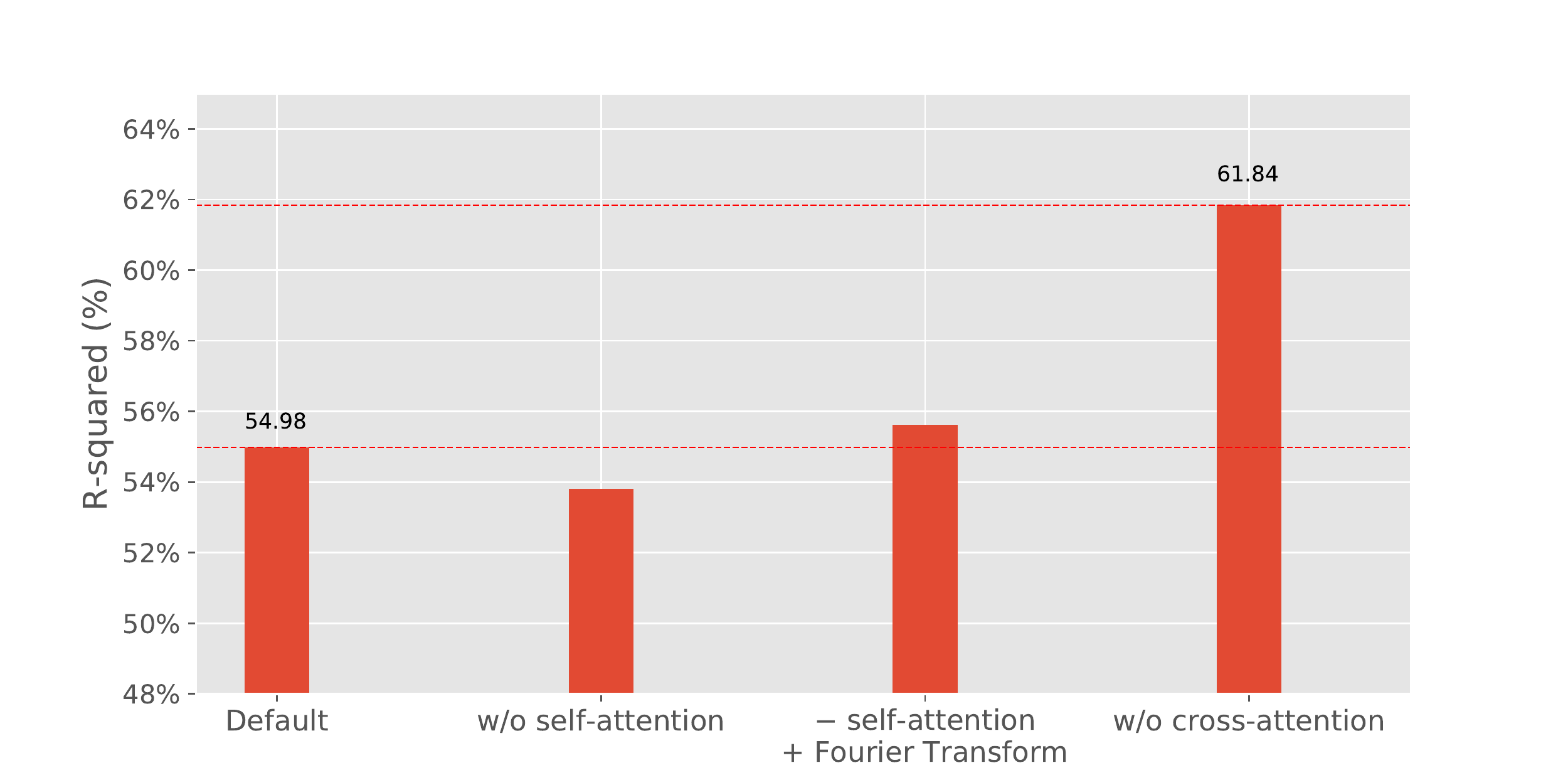}}\
\subfigure[Variants of Fourier-Net~\cite{LeeThorp2021FNetMT}.\label{fig:2b}]{\includegraphics[scale=0.4]{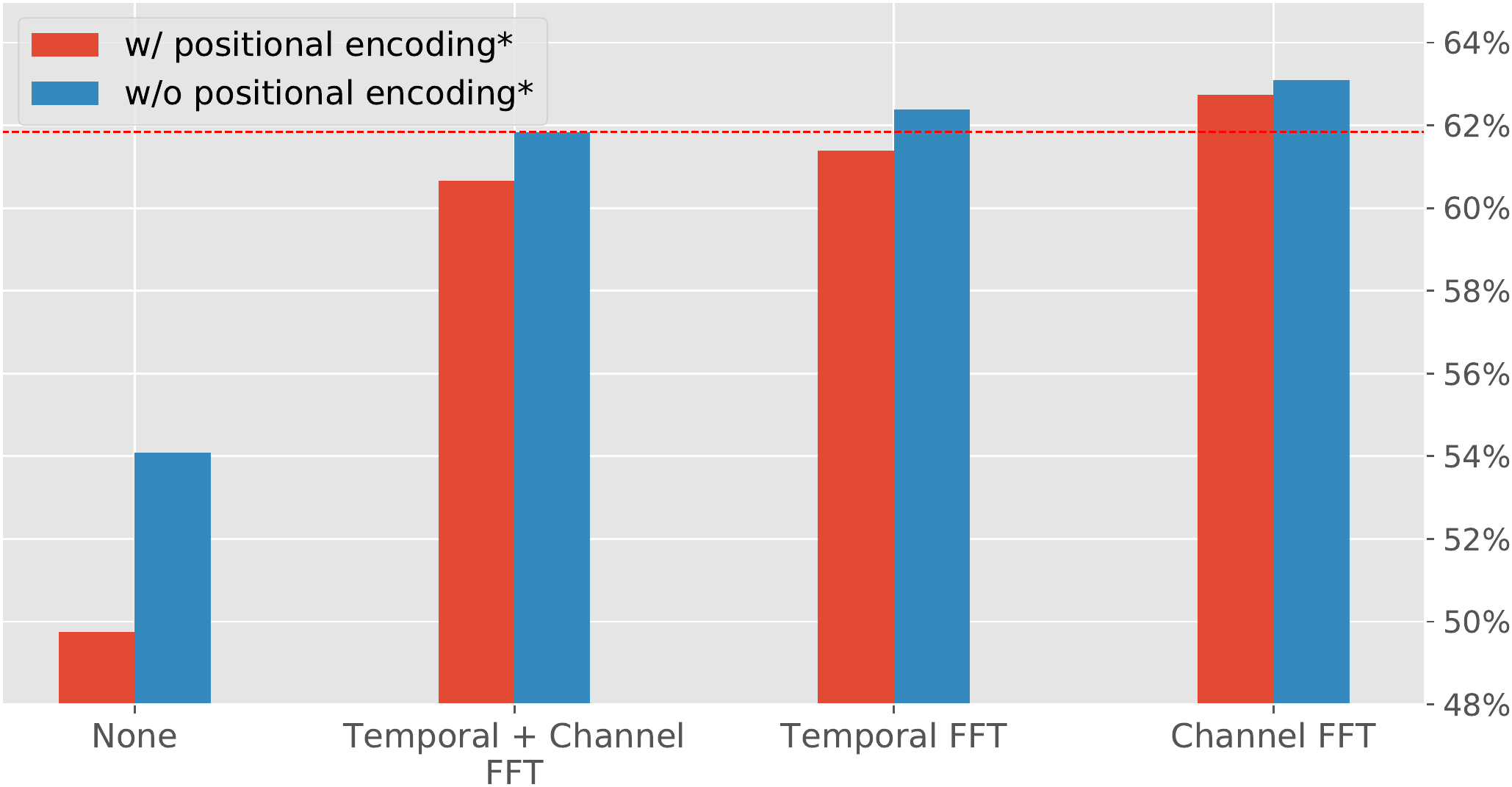}}
\vskip -0.1in
\caption{ETTh1~\cite{Zhou2021InformerBE} forecasting results of the modifications on Transformer and variants of Fourier-Net at 96-96 setting (The length of the historical horizon is set as 96 and the prediction length is 96). The higher R-squared score indicates better performance.}
\label{fig:2}
\end{figure*}

Although these Transformer-based models perform well on long-term time series forecasting tasks, there are still some problems to be addressed. First, there is a lack of explanation about the attention mechanism for capturing temporal dependency. A simple yet effective baseline DLinear~\cite{Zeng2022AreTE} questioned whether Transformer-based models are effective for time series forecasting. Second, Transformer-based models heavily rely on additional positional or date-specific encoding to guarantee the ordering of attention score, which may disturb the capture of temporal interaction. Third, while existing Transformer-based methods almost concentrate on how to reduce the complexity of attention computation, and have designed various sparse attention mechanisms through proper selection strategies to attain $O(L\log L)$ even $O(L)$ time and memory complexity where $L$ denotes the length of input sequence. However, these methods have a bulk of additional operations beyond attention, which makes the actual running time very long. To verify the effectiveness of the attention mechanism on time series forecasting, we conduct a group of experiments on ETTh1~\cite{Zhou2021InformerBE}. Figure~\ref{fig:2a} provides the forecasting results of the modifications on the Transformer. We can see that directly replacing the attention layer with Fourier Transform maintains the forecasting performance and removing the cross-attention significantly improves it. These results indicate that the attention mechanism on time series forecasting tasks may not be that effective. Figure~\ref{fig:2b} shows that applying a simple Fourier Transform can achieve the same, even better forecasting performance compared with attention-based models without extra positional or date-specific encoding. Additionally, individually capturing temporal and channel dependencies may bring extra improvement. 

\begin{figure}[h]
\vskip 0.2in
\centering
\subfigure[The redundancy of temporal information.]{\includegraphics[scale=0.4]{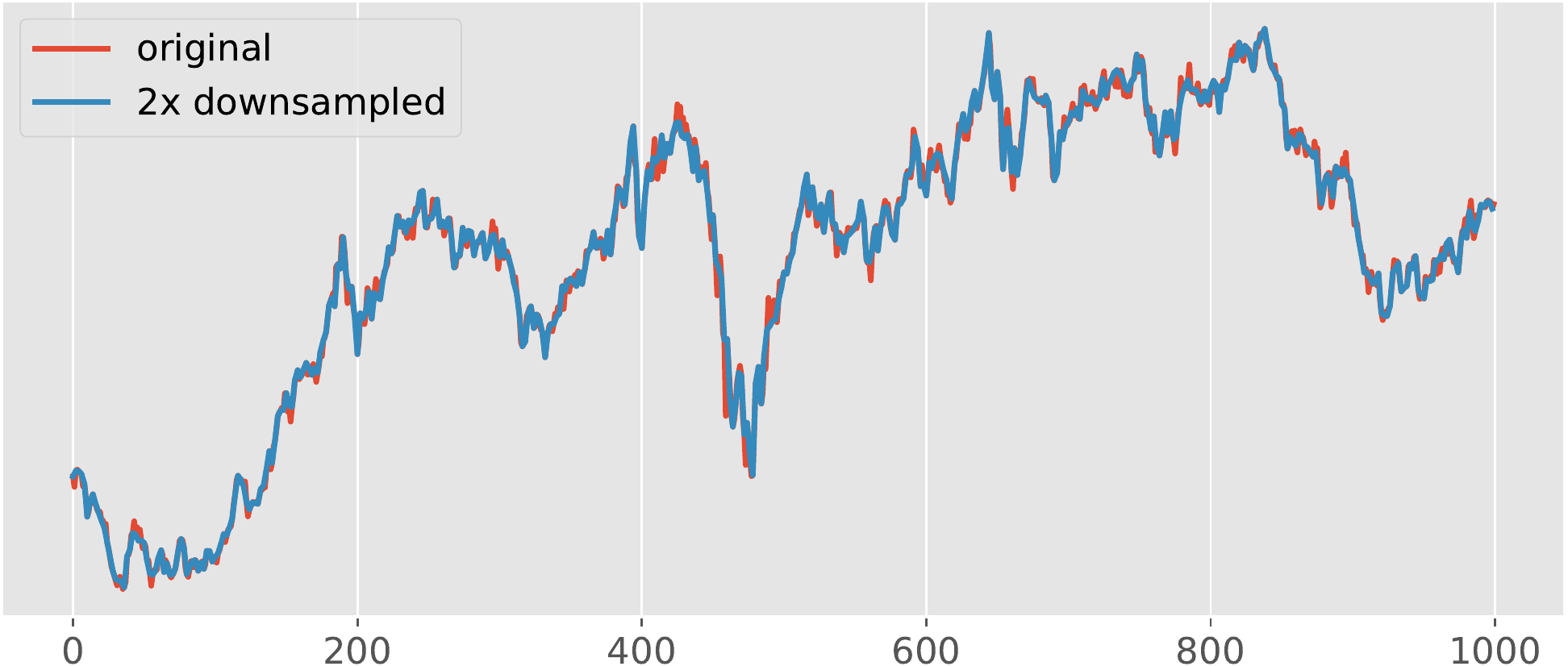}}\
\subfigure[The redundancy across different channels.]{\includegraphics[scale=0.4]{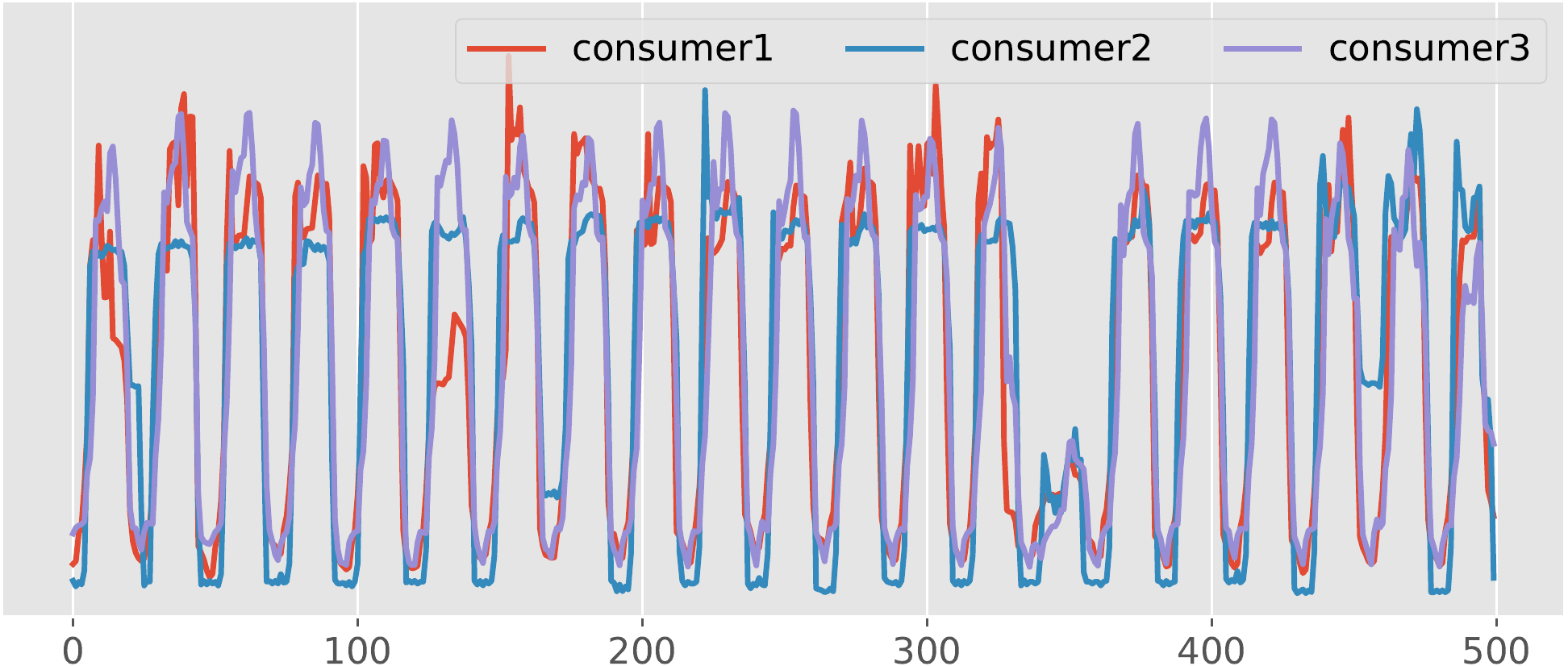}}
\vskip -0.1in
\caption{The redundancy of existing multivariate time series data. \textbf{top}: Exchange rate under different sampling rates. \textbf{bottom}: Electricity consumption of three consumers.}
\label{fig:3}
\end{figure}

Moreover, due to the difference in the sampling rate and the number of sensors, the current multivariate time series data from different scenarios often vary greatly in data form and have serious redundancy. Consider one instance of collected time series data as a tensor $\mathcal{X}\in\mathbb{R}^{n\times c}$, where $n$ denotes the length of $\mathcal{X}$ and $c$ is the dimension size. It is often uncertain which one is bigger or smaller between $n$ and $c$. That is, $\mathcal{X}$ generally has the low-rank property, such that $\text{rank}(\mathcal{X})\ll\min(n,c)$. Figure~\ref{fig:3} illustrates the ubiquitous redundancy of temporal and channel information. Specifically, the redundancy in temporal information is reflected in that the original sequence and the down-sampled sequence almost maintain the same temporal characteristics. The redundancy across channels occurs in that the information described by each channel may be consistent. Motivated by the above observations, we propose a novel and general framework for multivariate time series forecasting, named MTS-Mixers. The main contributions of this paper are:
\begin{itemize}
    \item We investigate the attention mechanism in time series forecasting and propose a novel and general framework named MTS-Mixers, which respectively capture temporal and channel dependencies.
    \item We leverage the low-rank property of existing time series data via factorized temporal and channel mixing and attain better prediction accuracy and efficiency.
    \item MTS-Mixers has achieved state-of-the-art forecasting performance on several public real-world multivariate time series datasets, yielding a 15.4\% MSE and a 12.9\% MAE reduction on average.
\end{itemize}

\section{Related Work}
Due to the ability of the attention mechanism to capture long-range dependencies, Transformer-based models have been widely used in language and vision tasks. \citet{Song2018AttendAD,Ma2019CDSACS,LI2019EnhancingTL} tried to directly apply vanilla Transformer to time series data but failed in long sequence forecasting tasks as self-attention operation scales quadratically with the input sequence length. \citet{Child2019GeneratingLS} applied sparse attention to reducing time complexity and memory usage for processing longer sequences. \citet{Zhou2021InformerBE} introduced non-autoregressive decoding to generate future time series to be predicted in one pass and designed selective attention for higher efficiency. \citet{Liu2022PyraformerLP} proposed tree structure pyramidal attention for lower complexity. To enhance the capture of temporal dependency, \citet{Wu2021AutoformerDT, Woo2022ETSformerES, Zhou2022FEDformerFE} disentangled time series data into trend and seasonality parts by removing the moving average and introduced frequency domain transformation. Recently, a simple but effective baseline~\cite{Zeng2022AreTE} questioned whether Transformers are effective for time series forecasting, reminding us to rethink the role of the attention mechanism.

Apart from the dot product attention mechanism, it is shown in Synthesizer~\cite{Tay2020SynthesizerRS}, MLP-Mixer~\cite{Tolstikhin2021MLPMixerAA}, FNet~\cite{LeeThorp2021FNetMT} and Poolformer~\cite{Diao2022MetaFormerAU} that, by replacing attention in Transformer with spatial MLP, Fourier Transform and pooling layer, the resulting models will deliver competitive performance in machine translation and computer vision domain. Unlike mentioned works above, we first investigate what we actually learn from attention-like modules in time series forecasting tasks, and then propose MTS-Mixers with factorized temporal and channel mixing to fit the inherent low-rank property of time series data.

\section{Preliminary}
%\section{Method}
\subsection{Problem definition}
%Let $\mathcal{X}=(\boldsymbol{x}_1,\boldsymbol{x}_2,\dots,\boldsymbol{x}_T)\in\mathbb{R}^{T\times c}$ be a multivariate time series instance with the length of $T$ , where $c$ is the dimension of each signal. 

Given a historical multivariate time series instance $\mathcal{X}_h=[\boldsymbol{x}_1,\boldsymbol{x}_2,\dots,\boldsymbol{x}_n]\in\mathbb{R}^{n\times c}$ with the length of $n$, time series forecasting tasks aim to predict the next $m$ steps values $\mathcal{X}_f=[\boldsymbol{x}_{n+1},\boldsymbol{x}_{n+2},\dots,\boldsymbol{x}_{n+m}]\in\mathbb{R}^{m\times c}$ across all the $c$ channels. Forecasting tasks are required to learn a map $\mathcal{F}:\mathcal{X}_h^{n\times c}\mapsto\mathcal{X}_f^{m\times c}$ where $\mathcal{X}_h$ and $\mathcal{X}_f$ are consecutive.

\subsection{Rethinking the mechanism of attention in Transformer-based forecasting models}\label{sec:3.2}

The general workflow of existing Transformer-based methods on time series forecasting tasks is illustrated in Figure~\ref{fig:1}. First, we generally utilize a 1D convolutional layer to obtain the input embedding $\mathcal{X}_\text{in}\in\mathbb{R}^{n\times d}$ from the original time series $\mathcal{X}_h\in\mathbb{R}^{n\times c}$ with the positional or date-specific encoding. Then on the encoder side, self-attention or other correlation patterns are used to capture token-level temporal similarity as
\begin{equation}
    \Tilde{\mathcal{X}}=\text{softmax}(\frac{\mathcal{X}_\text{in}\cdot\mathcal{X}_\text{in}^\top}{\sqrt{d}})\cdot\mathcal{X}=R_1\cdot\mathcal{X}_\text{in},
\end{equation}
where $R_1\in\mathbb{R}^{n\times n}$ describes token-wise temporal information. A feedforward neural network with two linear layers and activation function will be applied on $\Tilde{\mathcal{X}}$ to learn channel-wise features. On the decoder side, an initialized query $Q\in\mathbb{R}^{m\times d}$ is used to generate forecasting results as
\begin{equation}
    \Tilde{\mathcal{X}_f}=\text{softmax}(\frac{Q\cdot\Tilde{\mathcal{X}}^\top}{\sqrt{d}})\cdot\Tilde{\mathcal{X}}=R_2\cdot\Tilde{\mathcal{X}},
\end{equation}
where $R_2\in\mathbb{R}^{m\times n}$ describes the relationship between the input historical time series and the output prediction series. A projection layer is applied on  $\Tilde{\mathcal{X}_f}$ to obtain the final forecasting target $\mathcal{X}_f\in\mathbb{R}^{m\times c}$.

In essence, Transformer-based methods for time series forecasting contain two stages: learning token-wise temporal dependency across channels, and learning a map between input time series and output forecasting results. However, as shown in Figure~\ref{fig:2}, removing self-attention or cross-attention can maintain or even improve the forecasting performance. Replacing self-attention with Fourier Transform also delivers similar results, and respectively modeling temporal and channel dependencies will further enhance it. This means what we learn from attention-like modules is temporal dependency across all channels, and cross-attention bridges the relationship between input and output sequences. Attention mechanisms may not be necessary for capturing temporal dependency. Learning the mapping between input and output sequences and disentangling the modeling of temporal and channel dependencies may lead to better forecasting performance.

\begin{figure*}[ht]
\vskip 0.2in
\begin{center}
\centerline{\includegraphics[width=2\columnwidth]{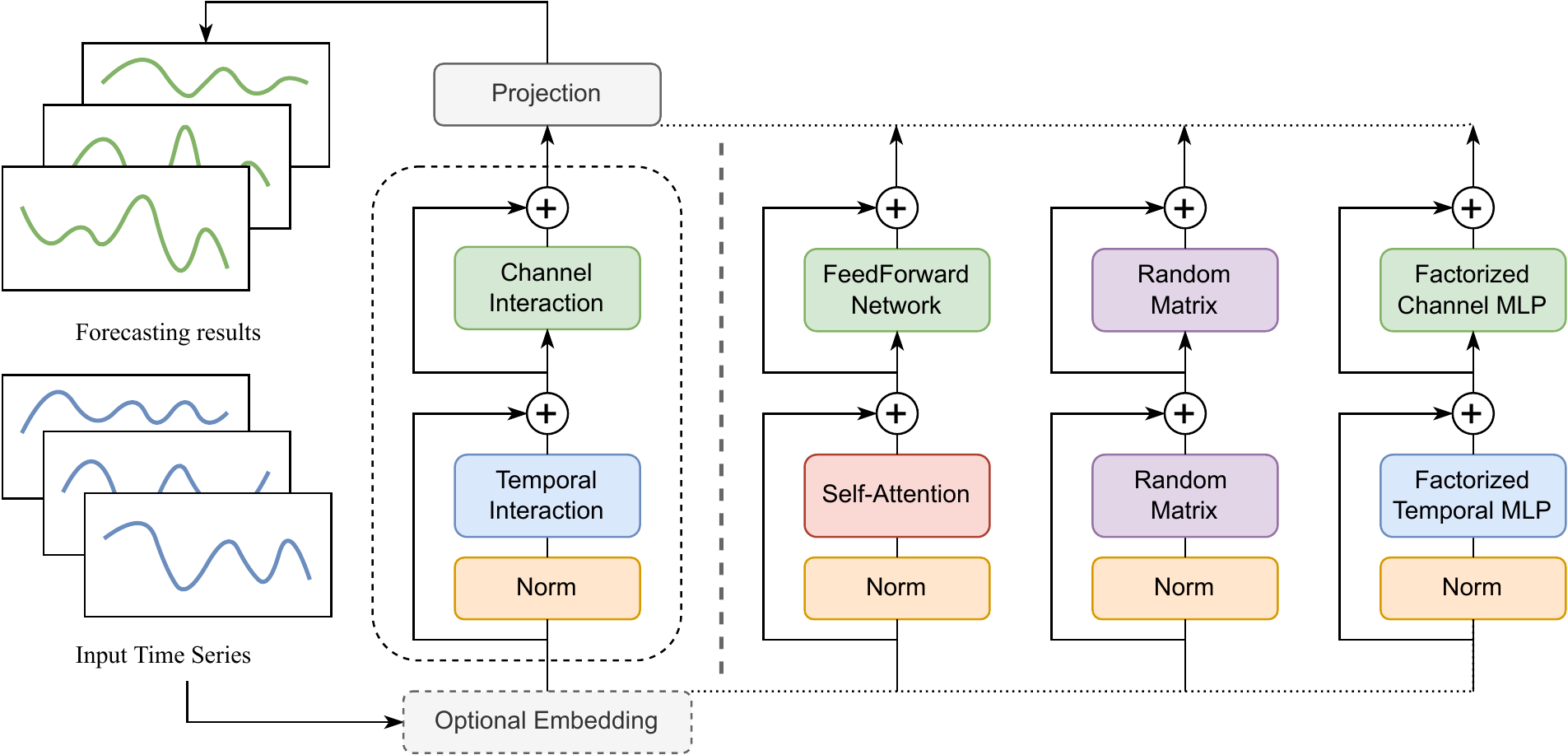}}
\caption{The overall architecture of MTS-Mixers. \textbf{Left}: the modules in the dashed box describe the general framework. \textbf{Right}: three specific implementations, where we can use attention, random matrix, or factorized MLP to capture dependencies.}
\label{fig:4}
\end{center}
\vskip -0.2in
\end{figure*}

\section{MTS-Mixers}\label{sec:4}
This section introduces MTS-Mixers, our proposed general framework for time series forecasting that respectively captures temporal and channel dependencies. The overall architecture of MTS-Mixers is shown in Figure~\ref{fig:4}, which aims to learn a map between the input $\mathcal{X}_h\in\mathbb{R}^{n\times c}$ and the output $\mathcal{X}_f\in\mathbb{R}^{m\times c}$. Notably, the input embedding module is optional. For better understanding, filters such as convolutional layers and the attention are simplified as a module for capturing temporal interaction. Then the other module for catching channel interaction is also applied, and the linear projection layer outputs the final forecasting results. Equation~(\ref{eq3}) depicts the above simplified time series forecasting process.
\begin{equation}\label{eq3}
\begin{aligned}
    \mathcal{X}_h^\mathcal{T}&=\texttt{Temporal}(\texttt{norm}(\mathcal{X}_h)), \\
    \mathcal{X}_h^\mathcal{C}&=\texttt{Channel}(\mathcal{X}_h+\mathcal{X}_h^\mathcal{T}), \\
    \mathcal{X}_f&=\texttt{Linear}(\mathcal{X}_h^\mathcal{T}+\mathcal{X}_h^\mathcal{C}).
\end{aligned}
\end{equation}
In this paper, we provide three specific implementations of MTS-Mixers. More details of them are as follows.

\textbf{Attention-based MTS-Mixer.} As Equation~(\ref{eq4}) shows, we add the sinusoidal positional encoding to obtain the input embedding $\Tilde{\mathcal{X}_h}$. Then, the multi-head self-attention will be used to capture temporal dependency $\mathcal{X}_h^\mathcal{T}$. A feed-forward network is to learn the channel interaction $\mathcal{X}_h^\mathcal{C}$. Compared with the vanilla Transformer, we remove its decoder and use one linear layer to directly learn the map between the learned features and the prediction sequence.
\begin{equation}\label{eq4}
\begin{aligned}
    \Tilde{\mathcal{X}_h}&=\texttt{norm}(\mathcal{X}_h)+\texttt{PE}(\mathcal{X}_h), \\
    \mathcal{X}_h^\mathcal{T}&=\texttt{Attention}(\Tilde{\mathcal{X}_h},\Tilde{\mathcal{X}_h},\Tilde{\mathcal{X}_h}), \\
    \mathcal{X}_h^\mathcal{C}&=\texttt{FFN}(\Tilde{\mathcal{X}_h}+\mathcal{X}_h^\mathcal{T}).
\end{aligned}
\end{equation}

\textbf{Random matrix MTS-Mixer.} As mentioned in Section~\ref{sec:3.2}, what we need to learn from the input $\mathcal{X}_h$ is three matrices: $T\in\mathbb{R}^{n\times n}$ describes the temporal dependency, $C\in\mathbb{R}^{c\times c}$ describes the channel dependency and the projection $F\in\mathbb{R}^{m\times n}$ with the optional bias or activation. Equation~(\ref{eq5}) demonstrates a time series forecasting process via matrix multiplication as
\begin{equation}\label{eq5}
    \mathcal{X}_f=F\cdot\sigma(T)\cdot\mathcal{X}_h\cdot\phi(C),
\end{equation}
where the operators $\sigma$ and $\phi$ denote the optional added bias or activation. Because the initialization of the matrices $F$, $T$, and $C$ are controllable, we call it a random matrix MTS-Mixer. The experiments in Section~\ref{sec:5.3} analyze the impact of different methods of initialization.

\begin{table*}[ht]
    \centering
    \caption{Multivariate time series forecasting results. The length of the historical horizon is set as 36 for ILI and 96 for the others. The prediction lengths are \{24,36,48,60\} for ILI and \{96, 192, 336, 720\} for others. The best results are highlighted in \textbf{bold}.}
    \label{tab:1}
    \vskip 0.15in
    \begin{small}
    % \begin{tabular}{*{16}{c}}
    \setlength\tabcolsep{5.5pt}
    \begin{tabular}{c|c|cc|cc|cc|cc|cc|cc|cc}
    \toprule
    \multicolumn{2}{c|}{Method} & \multicolumn{2}{c|}{MTS-Mixers} & \multicolumn{2}{c|}{FEDformer} & \multicolumn{2}{c|}{DLinear} & \multicolumn{2}{c|}{SCINet} & 
    \multicolumn{2}{c|}{Pyraformer} & \multicolumn{2}{c|}{Autoformer} & \multicolumn{2}{c}{Informer} \\
    \midrule
    \multicolumn{2}{c|}{Metric} & MSE & MAE & MSE & MAE & MSE & MAE & MSE & MAE & MSE & MAE
    & MSE & MAE & MSE & MAE \\
    \midrule
    \multirow{5}{*}{\rotatebox{90}{ECL}}
     & 96 & \textbf{0.146} & \textbf{0.246} & \underline{0.193} & 0.308 & 0.199 & \underline{0.284} & 0.205 & 0.312 & 0.386 & 0.449 & 0.201 & 0.317 & 0.274 & 0.368 \\
    & 192 & \textbf{0.163} & \textbf{0.260} & 0.201 & 0.315 & 0.198 & \underline{0.287} & \underline{0.197} & 0.308 & 0.378 & 0.443 & 0.222 & 0.334 & 0.296 & 0.386 \\
    & 336 & \textbf{0.175} & \textbf{0.273} & 0.214 & 0.329 & 0.210 & \underline{0.302} & \underline{0.202} & 0.312 & 0.376 & 0.443 & 0.231 & 0.338 & 0.300 & 0.394 \\
    & 720 & \textbf{0.202} & \textbf{0.295} & 0.246 & 0.355 & 0.245 & \underline{0.335} & \underline{0.234} & 0.338 & 0.376 & 0.445 & 0.254 & 0.361 & 0.373 & 0.439 \\
    \cmidrule{2-16}
    & Avg. & \textbf{0.172} & \textbf{0.269} & 0.214 & 0.327 & 0.213 & \underline{0.302} & \underline{0.210} & 0.318 & 0.379 & 0.445 & 0.227 & 0.338 & 0.311 & 0.397 \\
    \midrule
    \multirow{5}{*}{\rotatebox{90}{Traffic}}
     & 96 & \textbf{0.516} & \textbf{0.339} & \underline{0.587} & \underline{0.366} & 0.650 & 0.396 & 0.651 & 0.393 & 0.867 & 0.468 & 0.613 & 0.388 & 0.719 & 0.391 \\
    & 192 & \textbf{0.521} & \textbf{0.353} & \underline{0.604} & \underline{0.373} & 0.605 & 0.378 & 0.604 & 0.372 & 0.869 & 0.467 & 0.616 & 0.382 & 0.696 & 0.379 \\
    & 336 & \textbf{0.557} & \textbf{0.358} & 0.621 & 0.383 & \underline{0.612} & \underline{0.382} & 0.611 & 0.375 & 0.881 & 0.469 & 0.622 & 0.387 & 0.777 & 0.420 \\
    & 720 & \textbf{0.578} & \textbf{0.369} & \underline{0.626} & \underline{0.382} & 0.645 & 0.394 & 0.649 & 0.393 & 0.896 & 0.473 & 0.660 & 0.408 & 0.864 & 0.472 \\
    \cmidrule{2-16}
    & Avg. & \textbf{0.543} & \textbf{0.355} & \underline{0.609} & \underline{0.376} & 0.628 & 0.388 & 0.629 & 0.383 & 0.878 & 0.469 & 0.628 & 0.391 & 0.764 & 0.415 \\
    \midrule
    \multirow{5}{*}{\rotatebox{90}{PeMS04}}
     & 96 & \textbf{0.349} & \textbf{0.309} & \underline{0.427} & \underline{0.382} & 0.725 & 0.560 & 0.554 & 0.410 & 0.621 & 0.469 & 1.183 & 0.772 & 0.578 & 0.440 \\
    & 192 & \textbf{0.378} & \textbf{0.333} & \underline{0.449} & \underline{0.396} & 0.749 & 0.577 & 0.629 & 0.446 & 0.658 & 0.491 & 1.217 & 0.799 & 0.631 & 0.466 \\
    & 336 & \textbf{0.387} & \textbf{0.343} & \underline{0.464} & \underline{0.402} & 0.680 & 0.528 & 0.611 & 0.442 & 0.626 & 0.463 & 1.587 & 0.934 & 0.666 & 0.489 \\
    & 720 & \textbf{0.431} & \textbf{0.363} & \underline{0.521} & \underline{0.446} & 0.750 & 0.568 & 0.713 & 0.496 & 0.711 & 0.510 & 1.696 & 0.969 & 0.752 & 0.536 \\
    \cmidrule{2-16}
    & Avg. & \textbf{0.386} & \textbf{0.337} & \underline{0.465} & \underline{0.407} & 0.726 & 0.558 & 0.627 & 0.449 & 0.654 & 0.483 & 1.421 & 0.869 & 0.657 & 0.483 \\
    \midrule
    \multirow{5}{*}{\rotatebox{90}{Exchange}}
     & 96 & \textbf{0.079} & \textbf{0.197} & 0.148 & 0.278 & \underline{0.088} & \underline{0.218} & 0.142 & 0.249 & 1.748 & 1.105 & 0.197 & 0.323 & 0.847 & 0.752 \\
    & 192 & \textbf{0.172} & \textbf{0.295} & 0.271 & 0.380 & \underline{0.176} & \underline{0.315} & 0.261 & 0.364 & 1.874 & 1.151 & 0.300 & 0.369 & 1.204 & 0.895 \\
    & 336 & \underline{0.321} & \textbf{0.409} & 0.460 & 0.500 & \textbf{0.313} & \underline{0.427} & 0.457 & 0.490 & 1.943 & 1.172 & 0.509 & 0.524 & 1.672 & 1.036 \\
    & 720 & \underline{0.842} & \textbf{0.690} & 1.195 & 0.841 & \textbf{0.839} & \underline{0.695} & 1.364 & 0.859 & 2.085 & 1.206 & 1.447 & 0.941 & 2.478 & 1.310 \\
    \cmidrule{2-16}
    & Avg. & \textbf{0.354} & \textbf{0.398} & 0.519 & 0.500 & \textbf{0.354} & \underline{0.414} & 0.556 & 0.491 & 1.913 & 1.159 & 0.613 & 0.539 & 1.550 & 0.998 \\
    \midrule
    \multirow{5}{*}{\rotatebox{90}{Weather}}
     & 96 & \textbf{0.162} & \textbf{0.207} & 0.217 & 0.296 & \underline{0.196} & \underline{0.255} & 0.239 & 0.271 & 0.622 & 0.556 & 0.266 & 0.336 & 0.300 & 0.384 \\
    & 192 & \textbf{0.208} & \textbf{0.250} & 0.276 & 0.336 & \underline{0.237} & \underline{0.296} & 0.283 & 0.303 & 0.739 & 0.624 & 0.307 & 0.367 & 0.598 & 0.544 \\
    & 336 & \textbf{0.268} & \textbf{0.294} & 0.339 & 0.380 & \underline{0.283} & \underline{0.335} & 0.330 & 0.335 & 1.004 & 0.753 & 0.359 & 0.395 & 0.578 & 0.523 \\
    & 720 & \textbf{0.346} & \textbf{0.345} & 0.403 & 0.428 & \underline{0.347} & \underline{0.383} & 0.400 & 0.379 & 1.420 & 0.934 & 0.419 & 0.428 & 1.059 & 0.741 \\
    \cmidrule{2-16}
    & Avg. & \textbf{0.246} & \textbf{0.274} & 0.309 & 0.360 & \underline{0.266} & \underline{0.317} & 0.313 & 0.322 & 0.946 & 0.717 & 0.338 & 0.382 & 0.634 & 0.548 \\
    \midrule
    \multirow{5}{*}{\rotatebox{90}{ILI}}
    & 24 & \textbf{1.677} & \textbf{0.799} & 3.228 & 1.260 & \underline{2.398} & \underline{1.040} & 2.782 & 1.106 & 7.394 & 2.012 & 3.483 & 1.287 & 5.764 & 1.677 \\
    & 36 & \textbf{1.470} & \textbf{0.743} & 2.679 & 1.080 & \underline{2.646} & 1.088 & 2.689 & \underline{1.064} & 7.551 & 2.031 & 3.103 & 1.148 & 4.755 & 1.467 \\
    & 48 & \textbf{1.406} & \textbf{0.757} & 2.622 & 1.078 & 2.614 & 1.086 & \underline{2.324} & \underline{0.999} & 7.662 & 2.057 & 2.669 & 1.085 & 4.763 & 1.469 \\
    & 60 & \textbf{1.827} & \textbf{0.862} & 2.857 & 1.157 & 2.804 & 1.146 & 2.802 & \underline{1.112} & 7.931 & 2.100 & \underline{2.770} & 1.125 & 5.264 & 1.564 \\
    \cmidrule{2-16}
    & Avg. & \textbf{1.595} & \textbf{0.790} & 2.846 & 1.144 & \underline{2.616} & 1.090 & 2.649 & \underline{1.070} & 7.635 & 2.050 & 3.006 & 1.161 & 5.136 & 1.544 \\
    \bottomrule
    \end{tabular}
    \end{small}
    \vskip -0.1in
\end{table*}

\textbf{Factorized temporal and channel mixing.} Given the low-rank property of time series data described in Section~\ref{sec:1}, we design the factorized temporal and channel mixing strategies to capture dependencies with less redundancy. Specifically for the time series data with the temporal redundancy, we extract the temporal dependencies as Equation~(\ref{eq6})
\begin{equation}\label{eq6}
\begin{aligned}
    \mathcal{X}_{h,1},\dots,\mathcal{X}_{h,s}&=\texttt{sampled}(\texttt{norm}(\mathcal{X}_h)), \\
    \mathcal{X}_{h,i}^\mathcal{T}&=\texttt{Temporal}(\mathcal{X}_{h,i})\quad i\in[1,s], \\
    \mathcal{X}_h^\mathcal{T}&=\texttt{merge}(\mathcal{X}_{h,1}^\mathcal{T},\dots,\mathcal{X}_{h,s}^\mathcal{T}),
\end{aligned}
\end{equation}
where we first equidistantly downsample the original time series into $s$ interleaved subsequences. Then we individually utilize one temporal feature extractor (e.g., MLP or attention) to learn temporal information of those subsequences and merge them in the original order. For the time series data with channel redundancy, we reduce the noise of tensors corresponding to the time series in channel dimension by matrix decomposition as Equation~(\ref{eq7})
\begin{equation}\label{eq7}
\begin{aligned}
    \Tilde{\mathcal{X}_h^\mathcal{C}}&=\mathcal{X}_h+\mathcal{X}_h^\mathcal{T}, \\
    \Tilde{\mathcal{X}_h^\mathcal{C}}&=\mathcal{X}_h^\mathcal{C}+N\approx UV+N, \\
\end{aligned}
\end{equation}
where $N\in\mathbb{R}^{n\times c}$ represents the noise and $\mathcal{X}_h^\mathcal{C}\in\mathbb{R}^{n\times c}$ refers to the channel dependency after denoising. $U\in\mathbb{R}^{n\times m}$ and $V\in\mathbb{R}^{m\times c}$ ($m<c$) denote factorized channel interaction. We can use traditional methods such as truncated SVD~\cite{Rust1998TruncatingTS} and NMF~\cite{Geng2021IsAB} to obtain $\mathcal{X}_h^\mathcal{C}$. Here we provide a simple but effective factorizing method as
\begin{equation}\label{eq8}
    \mathcal{X}_h^\mathcal{C}=\sigma(\Tilde{\mathcal{X}_h^\mathcal{C}}\cdot W_1^\top+b_1)\cdot W_2^\top+b_2,
\end{equation}
where $W_1\in\mathbb{R}^{m\times c}$, $W_2\in\mathbb{R}^{c\times m}$ and $\sigma$ is an activation function. Different from MLP-like architectures in the vision domain, here we emphasize the temporal and channel redundancy of time series data and thus propose our factorization strategies. Notably, all the variants of our proposed MTS-Mixers can be mixed in an additive fashion if necessary. In practice, we only adopt the MLP-based MTS-Mixers via factorized temporal and channel mixing for the comparison with other baselines on forecasting tasks. See Appendix~\ref{ap:3.1} for more implementation details.

\section{Experiments}
\subsection{Experimental setup}

\begin{table*}[ht]
    \centering
    \caption{The forecasting performance of different types of models on ECL dataset. The length of historical horizon is set as 96, and the MSE and MAE results are averaged from 4 different prediction lengths \{96, 192, 336, 720\}.}
    \vskip 0.15in
    \begin{small}
    \begin{tabular}{wc{0.15\linewidth}|wc{0.28\linewidth}|wc{0.22\linewidth}|wc{0.07\linewidth}wc{0.07\linewidth}}
        \toprule
        Backbone & Interaction capture & Low-rank property usage & MSE & MAE \\
        \midrule
        \multirow{11}{*}{MTS-Mixers} & None & $\times$ & 0.215 & 0.304 \\
        & $+$ Temporal MLP & $\times$ & 0.189 & 0.280 \\
        & $+$ Channel MLP & $\times$ & 0.185 & 0.285 \\
        & $\dagger$ Factorized MLP & $\surd$ & \textbf{0.172} & \textbf{0.269} \\
        \cmidrule{2-5}
        & Attention & $\times$ & 0.215 & 0.316 \\
        & $\dagger$ Factorized Attention & $\surd$ & 0.206 & 0.309 \\
        & $+$ Channel Factorization & $\surd$ & 0.198 & 0.302 \\
        \cmidrule{2-5}
        & Random Matrix & $\times$ & 0.201 & 0.287 \\
        & Identity Matrix & $\times$ & 0.201 & 0.286 \\
        & $\dagger$ Factorized Matrix & $\surd$ & 0.195 & 0.286 \\
        & $+$ Channel Factorization & $\surd$ & 0.192 & 0.283 \\
        \midrule
        \multirow{2}{*}{FEDformer} & Autocorrelation + DWT & $\surd$ & 0.214 & 0.327 \\
        & $+$ Channel Factorization & $\surd$ & 0.205 & 0.319 \\
        \midrule
        \multirow{2}{*}{SCINet} & Tree-Conv & $\surd$ & 0.210 & 0.318 \\
        & $+$ Channel Factorization & $\surd$ & 0.197 & 0.295 \\
        \bottomrule
    \end{tabular}
    \end{small}
    \vskip -0.1in
    \label{tab:2}
\end{table*}

\textbf{Datasets.} We conduct extensive experiments on several public real-world benchmarks, covering economics, energy, traffic, weather, and infectious disease forecasting scenarios. Here is a detailed description of the datasets. 
(1) % \footnote{https://archive.ics.uci.edu/ml/datasets/ElectricityLoadDiagrams20112014}
ECL\footnote{https://archive.ics.uci.edu/ml/datasets/} records the hourly electricity consumption of 321 customers from 2012 to 2014. 
(2) ETT (Electricity Transformer Temperature)~\citep{Zhou2021InformerBE} consists of the data collected from electricity transformers, recording six power load features and oil temperature.
(3) Traffic\footnote{http://pems.dot.ca.gov/} contains the change of hourly road occupancy rates measured by hundreds of sensors on San Francisco Bay area freeways, which is collected from the California Department of Transportation. 
(4) PeMS04~\cite{Chen2001FreewayPM} records the change of traffic flow at 307 sensors in the Bay Area over 2 months from Jan 1st, 2018 to Feb 28th, 2018.
(5) Weather\footnote{https://www.ncei.noaa.gov/data/local-climatological-data/} contains 21 meteorological indicators like humidity and pressure in the 2020 year from nearly 1600 locations in the U.S.  
(6) Exchange~\citep{Lai2018ModelingLA} is a collection of exchange rates among eight different countries from 1990 to 2016.
(7) ILI\footnote{https://gis.cdc.gov/grasp/fluview/fluportaldashboard.html} records the weekly influenza-like illness (ILI) patients data from Centers for Disease Control and Prevention of the United States between 2002 and 2021, describing the ratio of patients observed with ILI and the total number of patients.
For a fair comparison, we follow the same standard protocol and split all forecasting datasets into training, validation, and test sets by the ratio of 6:2:2 for the ETT dataset and 7:1:2 for the other datasets.

\textbf{Baselines.} We select six baseline methods for comparison, including four latest state-of-the-art Transformer-based models: Informer~\cite{Zhou2021InformerBE}, Autoformer~\cite{Wu2021AutoformerDT}, Pyraformer~\cite{Liu2022PyraformerLP}, FEDformer~\cite{Zhou2022FEDformerFE} and two latest non-Transformer models SCINet~\cite{liu2022scinet} and DLinear~\cite{Zeng2022AreTE}. All the baselines follow the same evaluation protocol for a fair comparison. See Appendix~\ref{ap:3.2} for more details.

\textbf{Implementation details.} Our method is trained with the L2 loss, using the Adam optimizer~\citep{Kingma2015AdamAM}. The training process is early stopped within 10 epochs. MSE $\frac{1}{n}\sum_{i=1}^n(\boldsymbol{y}-\boldsymbol{\hat{y}})^2$ and MAE $\frac{1}{n}\sum_{i=1}^n\vert\boldsymbol{y}-\boldsymbol{\hat{y}}\vert$ are adopted as evaluation metrics on all the benchmarks. All the models are implemented in PyTorch~\citep{Paszke2019PyTorchAI} and trained/tested on a single Nvidia A100 40GB GPU for three times. See Appendix~\ref{ap:3} for more experimental details and hyper-parameters setting.

\begin{figure*}[ht]
\vskip 0.2in
\begin{center}
\centerline{\includegraphics[width=2\columnwidth]{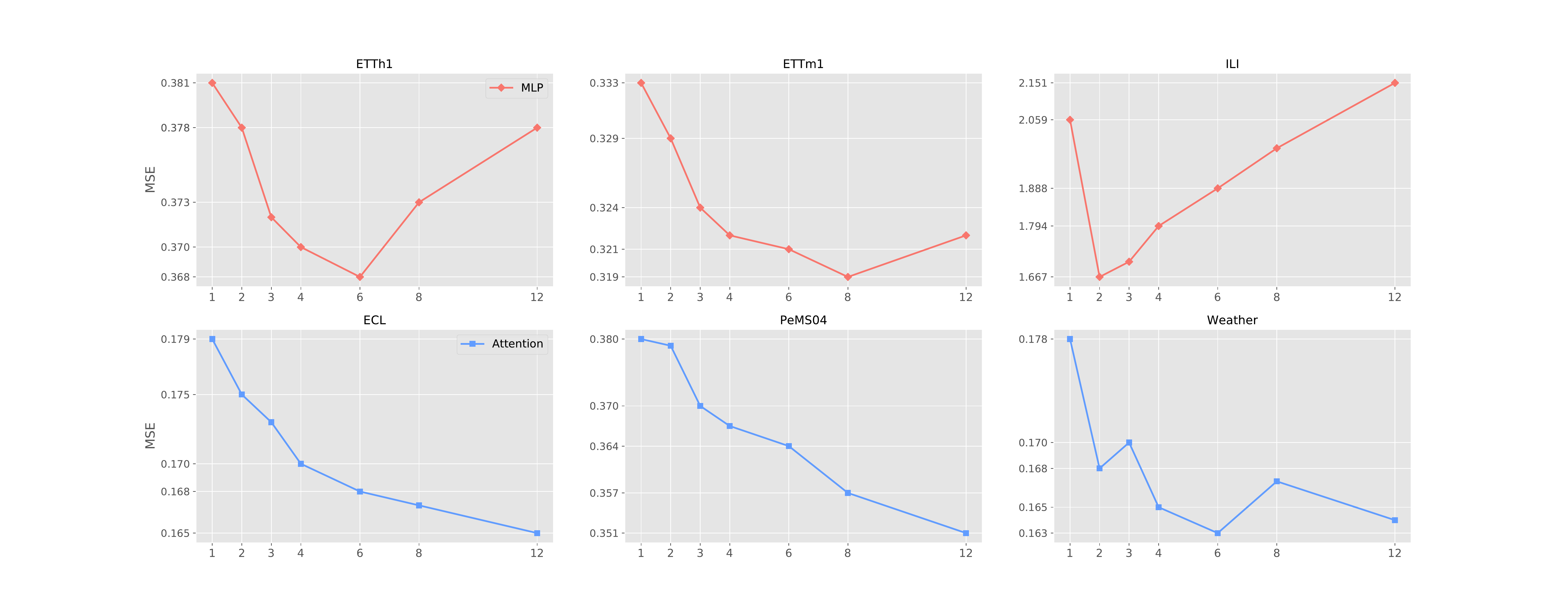}}
\caption{The impact of the hyper-parameter $s\in\{1,2,3,4,6,8,12\}$ which corresponds to the number of interleaved subsequences after downsampling the original time series data under the 96-96 setting. A lower MSE means better performance.}
\label{fig:5}
\end{center}
\vskip -0.2in
\end{figure*}

\subsection{Main results}
To fairly compare the forecasting performance, we follow the same evaluation protocol, where the length of the historical horizon is set as 36 for ILI and 96 for the others. The prediction lengths are \{24,36,48,60\} for ILI and \{96, 192, 336, 720\} for others. Table~\ref{tab:1} summarizes the results of multivariate time series forecasting on six datasets. The best results are highlighted in bold and the second best is underlined. MTS-Mixers achieves consistent state-of-the-art performance in all benchmarks. Especially compared with previous state-of-the-art results, MTS-Mixers yields a \textbf{18.1\%} averaged MSE reduction in ECL, \textbf{10.8\%} in Traffic, \textbf{17.0\%} in PeMS04, comparable results in Exchange, \textbf{7.2\%} in Weather and \textbf{39.0\%} in ILI. The full benchmarks on the ETT are provided in Appendix~\ref{ap:1}. See Appendix~\ref{ap:2} for more results of univariate time series forecasting. Note that the strong baseline FEDformer generally performs well on large datasets while DLinear performs well on small datasets. MTS-Mixers outperforms in all benchmarks and prediction length settings, implying its strength in various real-world scenarios.

\subsection{Ablation studies}\label{sec:5.3}
In this section, the ablation experiments are conducted, aiming at investigating: (1) the forecasting performance of the variants of MTS-Mixers; (2) the impact of factorized modules applied on other types of models, and (3) the detailed settings of temporal and channel factorization.

\textbf{Variants of MTS-Mixers.} 
Table~\ref{tab:2} summarizes the forecasting performance of the variants of MTS-Mixers and other types of models on the ECL dataset over four different prediction lengths. Notice that the "None" interaction capture refers to a simple linear layer, which has been also mentioned in~\cite{Zeng2022AreTE}. In general, our proposed MTS-Mixers based on MLP, attention, and random matrix all achieve better results than previous works, which verifies the effectiveness of this general framework on multivariate time series forecasting. All the variants of MTS-Mixers can gain consistent promotion via temporal and channel factorization, verifying that our proposed factorized strategies can be generalized to other models for capturing better temporal and channel interactions. Specifically, MTS-Mixers based on factorized MLP achieves the best prediction results, while the factorized attention outperforms the previous state-of-the-art Transformer-based models. MTS-Mixers based on factorized MLP does not need any positional or date-specific embedding, which guarantees the capture of temporal characteristics with less distortion. MTS-Mixers based on factorized attention outperforms vanilla attention with fewer parameters because the points in the subsequences after downsampling have more semantic information than before, which alleviates the over-fitting problem. Surprisingly, MTS-Mixers based on random matrix achieves better results than attention, which further explains the non-necessity of attention mechanism in the capture of the temporal dependency. Note that the "identity" means the parameters of the random matrix are initialized to the identity matrix, which indicates the initialization of the random matrix has slight effects on the forecasting performance.

Notably, some methods like FEDformer and SCINet also have tried to utilize the low-rand property of time series data. Specifically, FEDformer determines required features in the frequency domain through random selection to obtain sparser temporal information. SCINet also applies downsampling to the original time series, recursively generating odd and even subsequences in a tree structure to learn multi-scale temporal dependencies. We apply the channel factorization to them and also achieve a consistent promotion.

\begin{table*}[ht]
    \centering
    \caption{Running time (seconds) analysis for Transformer-based methods at different phases on the ECL dataset.}
    \vskip 0.15in
    \begin{small}
    \begin{tabular}{c|c|cccccc}
        \toprule
        \multirow{2}{*}{Phase} & \multirow{2}{*}{H} & MTS-Mixers & MTS-Mixers & \multirow{2}{*}{FEDformer} & \multirow{2}{*}{Autoformer} & \multirow{2}{*}{Informer} & \multirow{2}{*}{Transformer} \\ 
        & & (MLP) & (attention) & \\
        \midrule
        \multirow{4}{*}{Training} & 96 & 10.3 $\pm$ 0.4 & 12.1 $\pm$ 0.7 & 72.4 $\pm$ 4.8 & 29.1 $\pm$ 0.7 & 24.4 $\pm$ 0.7 & 20.6 $\pm$ 0.9 \\
        & 192 & 12.0 $\pm$ 0.4 & 14.5 $\pm$ 0.3 & 81.0 $\pm$ 3.6 & 31.4 $\pm$ 0.4 & 28.5 $\pm$ 1.1 & 22.8 $\pm$ 0.5 \\
        & 336 & 14.8 $\pm$ 0.3 & 17.2 $\pm$ 0.1 & 100.2 $\pm$ 3.1 & 38.9 $\pm$ 0.5 & 33.2 $\pm$ 0.2 & 27.5 $\pm$ 0.8 \\
        & 720 & 21.2 $\pm$ 0.6 & 23.0 $\pm$ 0.4 & 139.9 $\pm$ 1.0 & 65.0 $\pm$ 0.8 & 44.0 $\pm$ 1.1 & 39.0 $\pm$ 0.1 \\
        \midrule
        \multirow{4}{*}{Inference} & 96 & 10.3 $\pm$ 0.5 & 9.4 $\pm$ 1.3 & 13.8 $\pm$ 0.3 & 12.5 $\pm$ 0.2 & 10.7 $\pm$ 1.3 & 10.5 $\pm$ 1.2 \\
        & 192 & 12.8 $\pm$ 0.7 & 11.7 $\pm$ 0.4 & 18.6 $\pm$ 0.6 & 17.9 $\pm$ 1.3 & 16.1 $\pm$ 1.3 & 15.1 $\pm$ 1.1 \\
        & 336 & 16.6 $\pm$ 0.9 & 17.0 $\pm$ 0.8 & 26.0 $\pm$ 0.7 & 27.5 $\pm$ 0.3 & 21.3 $\pm$ 1.5 & 21.0 $\pm$ 1.7 \\
        & 720 & 26.2 $\pm$ 0.5 & 26.7 $\pm$ 0.6 & 44.9 $\pm$ 2.2 & 48.3 $\pm$ 2.2 & 40.1 $\pm$ 3.6 & 36.6 $\pm$ 0.2 \\
        \bottomrule
    \end{tabular}
    \end{small}
    \vskip -0.1in
    \label{tab:3}
\end{table*}

\textbf{Temporal factorization.} 
Figure~\ref{fig:5} demonstrates the impact of the hyper-parameter $s\in\{1,2,3,4,6,8,12\}$ which corresponds to the number of interleaved subsequences after downsampling the original time series data. Both MTS-Mixers based on MLP and attention achieve better forecasting performance after using temporal factorization. Notably, different datasets generally have different optimal $s$ because they are collected from various scenarios and sensors with particular sampling rates. MTS-Mixers based on attention can obtain more promotions with larger $s$ because it yields more semantic points with less redundancy. 

To further improve the efficiency of temporal factorization, we can use only one temporal feature extractor (i.e. shared parameters) for different downsampled subsequences. As Table~\ref{tab:4} shows, parameter-sharing temporal factorization can significantly reduce the parameter quantity with a trade-off.

\begin{table}[htbp]
\caption{The forecasting resutls of MTS-Mixers with shared parameters or not on the ETTm1 96-96 setting.}
\label{tab:4}
\vskip 0.15in
\begin{center}
\begin{small}
% \begin{sc}
\begin{tabular}{c|c|cc|c}
\toprule
Backbone & $s$ & MSE & MAE & params. \\
\midrule
MLP & 1 & 0.333 & 0.370 & 0.27M \\
MLP & 8 & 0.319 & 0.357 & 0.28M \\
MLP (shared) & 8 & 0.360 & 0.391 & 0.10M \\
\midrule
Attention & 1 & 0.358 & 0.389 & 4.23M \\
Attention & 8 & 0.347 & 0.382 & 33.77M \\
Attention (shared) & 8 & 0.348 & 0.383 & 4.23M \\
\bottomrule
\end{tabular}
% \end{sc}
\end{small}
\end{center}
\vskip -0.1in
\end{table}

\begin{table}[htbp]
\caption{The forecasting resutls of MTS-Mixers with different channel factorization strategies on the ECL 96-96 setting.}
\label{tab:5}
\vskip 0.15in
\begin{center}
\begin{small}
\begin{tabular}{c|cc}
\toprule
Decomposition & MSE & MAE \\
\midrule
None & 0.179 & 0.275 \\
Truncated SVD~\cite{Rust1998TruncatingTS} & 0.162 & 0.254 \\
NMF~\cite{Geng2021IsAB} & 0.159 & 0.251 \\
Channel Drop~\cite{Kong2021ReflashDI} & 0.164 & 0.256 \\
Factorized MLP (ours) & 0.146 & 0.246 \\
\bottomrule
\end{tabular}
\end{small}
\end{center}
\vskip -0.1in
\end{table}

\textbf{Channel factorization.} As mentioned in Section~\ref{sec:4}, here we study the effects of other decomposition methods on channel factorization. Table~\ref{tab:5} provides four denoising strategies, which all successfully alleviate the redundancy across channels and achieve higher prediction accuracy. Specifically, we only retain 10\% of the maximum singular value in truncated SVD and follow the same setting of NMF in~\citet{Geng2021IsAB}. As for channel drop, we randomly discard 10\% channels and set them as 0. The experiments prove the necessity of channel factorization in time series forecasting.

\subsection{Model analysis}

\begin{figure}[h]
\vskip 0.2in
\centering
\subfigure[Visual analysis of Transformer attention on ECL.]{\includegraphics[scale=0.34]{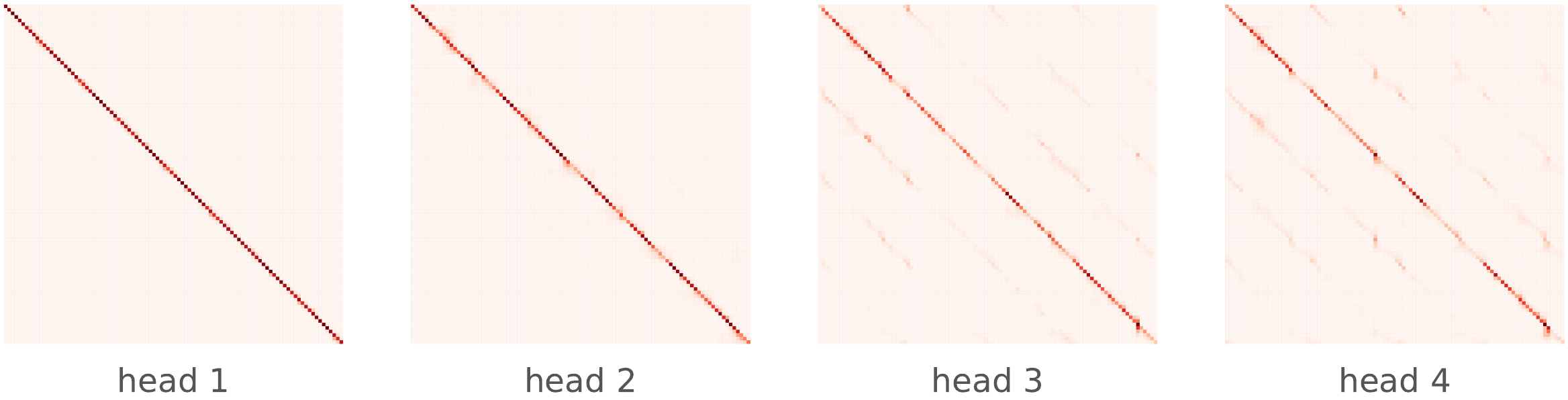}}\
\subfigure[Visual analysis of MTS-Mixers on ECL.]{\includegraphics[scale=0.34]{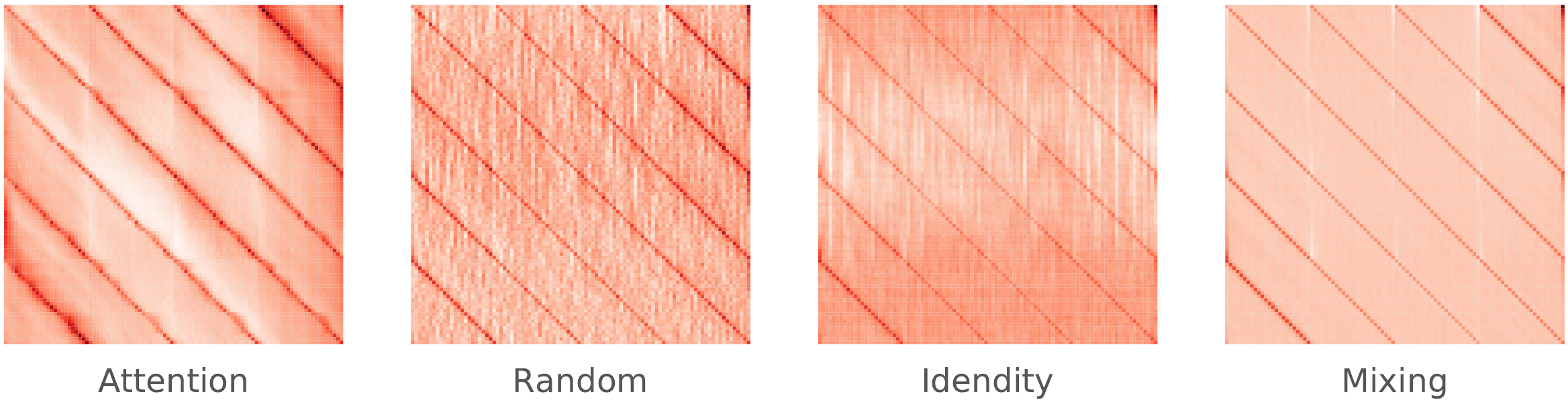}}
\vskip -0.1in
\caption{The visual analysis of Transformer and MTS-Mixers on the ECL. Only one layer of the encoder is used for each model.}
\label{fig:6}
\end{figure}

\textbf{Prediction patterns analysis.} As Figure~\ref{fig:6} shows, we investigate what patterns the models learn in forecasting tasks. Specifically, most of the heads in attention of vanilla Transformer may hardly capture any useful information. Instead, the variants of MTS-Mixers consistently learn the similar mapping matrix with periodic characteristics. We provide additional analysis and visualisation of weights for other datasets in Appendices~\ref{ap:4} and~\ref{ap:5}.

\textbf{Runtime analysis.} We conduct runtime analysis for Transformer based methods on the ECL dataset. Each of them is run three times for one epoch. As Table~\ref{tab:3} shows, our proposed MTS-Mixers achieve better efficiency in the training and inference stages. Although some previous works have emphasized their contribution to reducing the complexity of attention, their operational efficiency is not high due to a bulk of additional processing such as Wavelet Transform, reminding us to challenge their claimed efficiency.

\section{Conclusion}
This paper proposes MTS-Mixers, a general framework for multivariate time series forecasting. We conducted an extensive study to investigate the real contribution and deficiencies of attention mechanisms on the performance of time series forecasting. The experimental results have demonstrated that attention is unnecessary for capturing temporal dependencies, and the redundancy in time series data affects the forecasting performance. Furthermore, our proposed temporal and channel factorization strategies leverage the low-rank property of time series data and have achieved state-of-the-art results on several real-world datasets with higher efficiency. Model analysis on patterns that the models have learned indicates that the mapping between the input and output sequences may be the key we need.


\bibliography{main}
\bibliographystyle{icml2023}

%%%%%%%%%%%%%%%%%%%%%%%%%%%%%%%%%%%%%%%%%%%%%%%%%%%%%%%%%%%%%%%%%%%%%%%%%%%%%%%
%%%%%%%%%%%%%%%%%%%%%%%%%%%%%%%%%%%%%%%%%%%%%%%%%%%%%%%%%%%%%%%%%%%%%%%%%%%%%%%
% APPENDIX
%%%%%%%%%%%%%%%%%%%%%%%%%%%%%%%%%%%%%%%%%%%%%%%%%%%%%%%%%%%%%%%%%%%%%%%%%%%%%%%
%%%%%%%%%%%%%%%%%%%%%%%%%%%%%%%%%%%%%%%%%%%%%%%%%%%%%%%%%%%%%%%%%%%%%%%%%%%%%%%
\newpage
\appendix
\onecolumn
\section{Full Benchmark on the ETT Datasets}\label{ap:1}

\begin{table*}[htbp]
    \centering
    % \vskip -0.1in
    \caption{The full multivariate time series forecasting benchmarks of ETT datasets. The length of historical horizon is set as 96 and the prediction lengths are \{96, 192, 336, 720\} for all the datasets. The best results are highlighted in \textbf{bold}. The second best is \underline{underlined}.}
    \label{tab:6}
    \vskip 0.15in
    \begin{small}
    % \begin{tabular}{*{16}{c}}
    \setlength\tabcolsep{5.5pt}
    \begin{tabular}{c|c|cc|cc|cc|cc|cc|cc|cc}
    \toprule
    \multicolumn{2}{c|}{Method} & \multicolumn{2}{c|}{MTS-Mixers} & \multicolumn{2}{c|}{FEDformer} & \multicolumn{2}{c|}{DLinear} & \multicolumn{2}{c|}{SCINet} & 
    \multicolumn{2}{c|}{Pyraformer} & \multicolumn{2}{c|}{Autoformer} & \multicolumn{2}{c}{Informer} \\
    \midrule
    \multicolumn{2}{c|}{Metric} & MSE & MAE & MSE & MAE & MSE & MAE & MSE & MAE & MSE & MAE
    & MSE & MAE & MSE & MAE \\
    \midrule
    \multirow{5}{*}{\rotatebox{90}{ETTh1}}
     & 96 & \textbf{0.368} & \textbf{0.393} & \underline{0.376} & 0.419 & 0.386 & \underline{0.400} & 0.404 & 0.415 & 0.664 & 0.612 & 0.449 & 0.459 & 0.865 & 0.713 \\
    & 192 & \textbf{0.419} & \textbf{0.425} & \underline{0.420} & 0.448 & 0.437 & \underline{0.432} & 0.456 & 0.445 & 0.790 & 0.681 & 0.500 & 0.482 & 1.008 & 0.792 \\
    & 336 & \underline{0.466} & \textbf{0.453} & \textbf{0.459} & 0.465 & 0.481 & \underline{0.459} & 0.519 & 0.481 & 0.891 & 0.738 & 0.521 & 0.496 & 1.107 & 0.809 \\
    & 720 & \textbf{0.473} & \textbf{0.470} & \underline{0.506} & \underline{0.507} & 0.519 & 0.516 & 0.564 & 0.528 & 0.963 & 0.782 & 0.514 & 0.512 & 1.181 & 0.865 \\
    \cmidrule{2-16}
    & Avg. & \textbf{0.432} & \textbf{0.435} & \underline{0.440} & 0.460 & 0.456 & \underline{0.452} & 0.486 & 0.467 & 0.827 & 0.703 & 0.496 & 0.487 & 1.040 & 0.795 \\
    \midrule
    \multirow{5}{*}{\rotatebox{90}{ETTm1}}
     & 96 & \textbf{0.319} & \textbf{0.357} & 0.379 & 0.419 & \underline{0.345} & \underline{0.372} & 0.350 & 0.385 & 0.543 & 0.510 & 0.505 & 0.475 & 0.672 & 0.571 \\
    & 192 & \textbf{0.363} & \textbf{0.384} & 0.426 & 0.441 & \underline{0.380} & \underline{0.389} & 0.382 & 0.400 & 0.557 & 0.537 & 0.553 & 0.496 & 0.795 & 0.669 \\
    & 336 & \textbf{0.397} & \textbf{0.408} & 0.445 & 0.459 & \underline{0.413} & \underline{0.413} & 0.419 & 0.425 & 0.754 & 0.655 & 0.621 & 0.537 & 1.212 & 0.871 \\
    & 720 & \textbf{0.456} & \textbf{0.445} & 0.543 & 0.490 & \underline{0.474} & \underline{0.453} & 0.494 & 0.463 & 0.908 & 0.724 & 0.671 & 0.561 & 1.166 & 0.823 \\
    \cmidrule{2-16}
    & Avg. & \textbf{0.384} & \textbf{0.399} & 0.448 & 0.452 & \underline{0.403} & \underline{0.407} & 0.411 & 0.418 & 0.691 & 0.607 & 0.588 & 0.517 & 0.961 & 0.734 \\
    \midrule
    \multirow{5}{*}{\rotatebox{90}{ETTh2}}
     & 96 & \textbf{0.303} & \textbf{0.352} & 0.358 & 0.397 & 0.333 & 0.387 & \underline{0.312} & \underline{0.355} & 0.645 & 0.597 & 0.346 & 0.388 & 3.755 & 1.525 \\
    & 192 & \textbf{0.390} & \textbf{0.407} & 0.429 & 0.439 & 0.477 & 0.476 & \underline{0.401} & \underline{0.412} & 0.788 & 0.683 & 0.456 & 0.452 & 5.602 & 1.931 \\
    & 336 & \underline{0.436} & \underline{0.437} & 0.496 & 0.487 & 0.594 & 0.541 & \textbf{0.413} & \textbf{0.432} & 0.907 & 0.747 & 0.482 & 0.486 & 4.721 & 1.835 \\
    & 720 & \textbf{0.426} & \textbf{0.440} & \underline{0.463} & \underline{0.474} & 0.831 & 0.657 & 0.490 & 0.483 & 0.963 & 0.783 & 0.515 & 0.511 & 3.647 & 1.625 \\
    \cmidrule{2-16}
    & Avg. & \textbf{0.389} & \textbf{0.409} & 0.437 & 0.449 & 0.559 & 0.515 & \underline{0.404} & \underline{0.421} & 0.826 & 0.703 & 0.450 & 0.459 & 4.431 & 1.729 \\
    \midrule
    \multirow{5}{*}{\rotatebox{90}{ETTm2}}
     & 96 & \textbf{0.177} & \textbf{0.261} & 0.203 & 0.287 & \underline{0.193} & 0.292 & 0.201 & \underline{0.280} & 0.435 & 0.507 & 0.255 & 0.339 & 0.365 & 0.453 \\
    & 192 & \textbf{0.240} & \textbf{0.300} & \underline{0.269} & \underline{0.328} & 0.284 & 0.362 & 0.283 & 0.331 & 0.730 & 0.673 & 0.281 & 0.340 & 0.533 & 0.563 \\
    & 336 & \textbf{0.304} & \textbf{0.343} & 0.325 & 0.366 & 0.369 & 0.427 & \underline{0.318} & \underline{0.352} & 1.201 & 0.845 & 0.339 & 0.372 & 1.363 & 0.887 \\
    & 720 & \textbf{0.395} & \textbf{0.397} & \underline{0.421} & \underline{0.415} & 0.554 & 0.522 & 0.439 & 0.423 & 3.625 & 1.451 & 0.422 & 0.419 & 3.379 & 1.388 \\
    \cmidrule{2-16}
    & Avg. & \textbf{0.279} & \textbf{0.325} & \underline{0.304} & 0.349 & 0.350 & 0.401 & 0.310 & \underline{0.347} & 1.498 & 0.869 & 0.324 & 0.368 & 1.410 & 0.823 \\
    \bottomrule
    \end{tabular}
    \end{small}
    % \vskip -0.1in
\end{table*}

Table~\ref{tab:6} shows the full multivariate time series forecasting benchmarks of four ETT datasets, including the hourly recorded ETTh1 and ETTh2; minutely recorded ETTm1 and ETTm2. MTS-Mixers also achieve consistent state-of-the-art performance in four ETT benchmarks. Specifically, MTS-Mixers yields a \textbf{1.8\%} averaged MSE reduction in ETTh1, \textbf{4.7\%} in ETTm1, \textbf{3.7\%} in ETTh2 and \textbf{8.2\%} in ETTm2 compared with the second best results.

\section{Univariate Time Series Forecasting}\label{ap:2}

\begin{table*}[!hbp]
    \centering
    % \vskip -0.1in
    \caption{The univariate time series forecasting results. The length of historical horizon is set as 96 and the prediction lengths are \{96, 192, 336, 720\ for all the datasets. The best results are highlighted in \textbf{bold}. The second best is \underline{underlined}.}
    \label{tab:7}
    \vskip 0.15in
    \begin{small}
    % \begin{tabular}{*{16}{c}}
    \setlength\tabcolsep{5.5pt}
    \begin{tabular}{c|c|cc|cc|cc|cc|cc|cc|cc}
    \toprule
    \multicolumn{2}{c|}{Method} & \multicolumn{2}{c|}{MTS-Mixers} & \multicolumn{2}{c|}{DLinear} & \multicolumn{2}{c|}{FEDformer} & \multicolumn{2}{c|}{SCINet} & 
    \multicolumn{2}{c|}{N-HiTs} & \multicolumn{2}{c|}{N-BEATS} & \multicolumn{2}{c}{Autoformer} \\
    \midrule
    \multicolumn{2}{c|}{Metric} & MSE & MAE & MSE & MAE & MSE & MAE & MSE & MAE & MSE & MAE
    & MSE & MAE & MSE & MAE \\
    \midrule
    \multirow{5}{*}{\rotatebox{90}{Exchange}}
     & 96 & \textbf{0.096} & \textbf{0.226} & 0.110 & 0.264 & 0.140 & 0.295 & \underline{0.101} & \underline{0.238} & 0.114 & 0.248 & 0.156 & 0.299 & 0.153 & 0.306 \\
    & 192 & \textbf{0.199} & \textbf{0.335} & \underline{0.223} & 0.383 & 0.278 & 0.411 & \underline{0.223} & \underline{0.351} & 0.250 & 0.387 & 0.669 & 0.665 & 0.302 & 0.424 \\
    & 336 & \textbf{0.389} & \textbf{0.471} & \underline{0.405} & 0.509 & 0.515 & 0.554 & 0.429 & \underline{0.493} & 0.434 & 0.516 & 0.611 & 0.605 & 0.609 & 0.615 \\
    & 720 & 1.074 & 0.791 & \textbf{1.044} & \underline{0.782} & 1.319 & 0.891 & 1.094 & 0.796 & \underline{1.061} & \textbf{0.773} & 1.111 & 0.860 & 1.261 & 0.873 \\
    \cmidrule{2-16}
    & Avg. & \textbf{0.440} & \textbf{0.456} & \underline{0.446} & 0.485 & 0.563 & 0.538 & 0.462 & \underline{0.470} & 0.465 & 0.481 & 0.637 & 0.607 & 0.581 & 0.555 \\
    \midrule
    \multirow{5}{*}{\rotatebox{90}{ETTm2}}
     & 96 & \textbf{0.065} & \textbf{0.184} & 0.071 & \underline{0.194} & \underline{0.069} & 0.198 & 0.074 & 0.197 & 0.092 & 0.232 & 0.082 & 0.219 & 0.099 & 0.241 \\
    & 192 & \textbf{0.100} & \textbf{0.235} & 0.104 & 0.238 & \underline{0.103} & 0.246 & \underline{0.103} & \underline{0.237} & 0.128 & 0.276 & 0.120 & 0.268 & 0.132 & 0.276 \\
    & 336 & \textbf{0.131} & \textbf{0.276} & 0.143 & 0.288 & \underline{0.131} & \underline{0.281} & 0.135 & 0.280 & 0.165 & 0.314 & 0.226 & 0.370 & 0.154 & 0.305 \\
    & 720 & \textbf{0.184} & \textbf{0.334} & 0.192 & 0.336 & \underline{0.185} & \underline{0.331} & \underline{0.185} & 0.335 & 0.243 & 0.397 & 0.188 & 0.338 & 0.205 & 0.353 \\
    \cmidrule{2-16}
    & Avg. & \textbf{0.120} & \textbf{0.257} & 0.128 & 0.264 & \underline{0.122} & 0.264 & 0.124 & \underline{0.262} & 0.157 & 0.305 & 0.154 & 0.299 & 0.148 & 0.294 \\
    \bottomrule
    \end{tabular}
    \end{small}
    % \vskip -0.1in
\end{table*}

Table~\ref{tab:7} provides the univariate time series forecasting results on two typical datasets. Here we include two baselines N-HiTs~\cite{Challu2022NHiTSNH} and N-BEATS~\cite{Oreshkin2019NBEATSNB} which are widely used in univariate benchmarks. Notably, for univariate time series forecasting, we remove the channel factorization module in MTS-Mixers. MTS-Mixers yields a \textbf{1.3\%} averaged MSE reduction in Exchange and \textbf{1.6\%} in ETTm2 univariate benchmarks, proving the effectiveness of our proposed temporal factorization strategy.

\section{Experimental Details}\label{ap:3}
\subsection{Reproduction Details for MTS-Mixers}\label{ap:3.1}
As shown in Figure~\ref{fig:2}, we have proposed three specific implementations of MTS-Mixers. For attention-based MTS-Mixer, we do not use convolution operation to obtain tokens but directly capture temporal dependency on each channel via multi-head self-attention. The number of the head is set as the least prime factor of $c$ which is the number of channels of the time series instance $\mathcal{X}\in\mathbb{R}^{n\times c}$. The feedforward neural network (FFN) consists of two fully-connected layers and a GELU~\cite{Hendrycks2016GaussianEL} activation. The number of hidden states in FFN is selected from $\{64, 512, 2048\}$ which corresponds to $c$ of different datasets. For matrix-based MTS-Mixer, we provide two initialization forms, including random and identity matrix initialization. For MLP-based MTS-Mixer, the temporal and channel MLPs both contain two linear layers with a GELU activation. The number of hidden states is set as 512 for temporal MLP and $m$ for channel MLP. The number of stacked units in MTS-Mixers for capturing temporal and channel interaction is all set as 2 for a fair comparison. We adopt reversible instance normalization~\cite{Kim2022ReversibleIN} rather than disentanglement to alleviate the distribution shift problem.

\begin{figure}[ht]
\vskip 0.2in
\begin{center}
\centerline{\includegraphics[width=0.9\columnwidth]{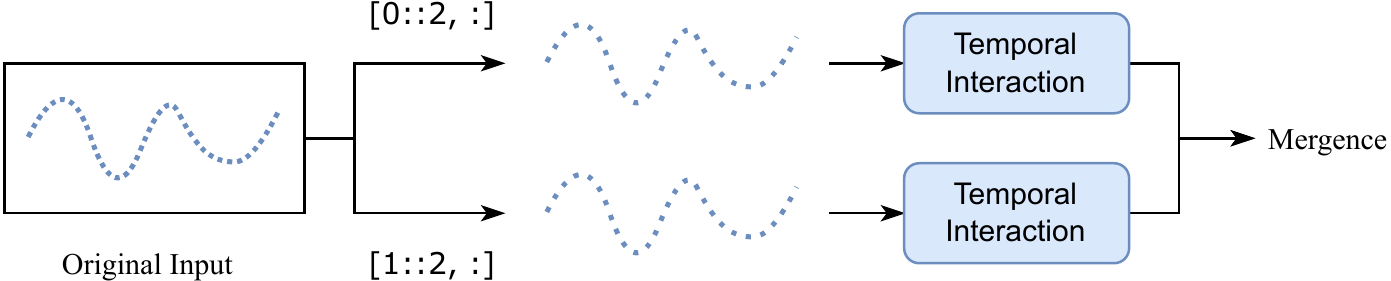}}
\caption{An example of temporal factorization when $s=2$.}
\label{fig:7}
\end{center}
\vskip -0.2in
\end{figure}

\textbf{Temporal factorization}. Figure~\ref{fig:7} illustrates an example of temporal factorization where we down-sample the original time series into two interleaved subsequences. In practice, We apply equidistant down-sampling in the original time series $\mathcal{X}\in\mathbb{R}^{n\times c}$ as
\begin{equation}
    \mathcal{X}_{h,i} = \Tilde{\mathcal{X}}_h[i-1::s, :],\quad 1\leq i\leq s,
\end{equation}
where $i$ denotes the $i$-th subsequence and $[\cdot]$ means a slice operation. We perform a grid search over the number of interleaved subsequences $s\in\{1, 2, 3, 4, 6, 8, 12\}$ on all benchmarks.

\textbf{Channel factorization}. As mentioned in Equation~(\ref{eq7}) and~(\ref{eq8}), we use a channel MLP to learn factorized channel interaction $\mathcal{X}_h^\mathcal{C}=UV$ where $U\in\mathbb{R}^{n\times m}$ and $V\in\mathbb{R}^{m\times c}$ ($m<c$). We perform a grid search over $m\in\{0, 16, 64\}$ (0 means no interaction among channels).

\subsection{Details on Benchmark Tasks and Baselines}\label{ap:3.2}
We conduct experiments on ten time series datasets and all of them follow the same pre-processing protocol in~\cite{Zhou2021InformerBE}. The random seed is set as 1024 in the training phase. Table~\ref{tab:8} provides the initial learning rate and batch size used in different datasets. For time series forecasting baselines, the results of FEDformer, DLinear, SCINet, Autoformer, and Informer are based on our reproduction. The hyper-parameters are the same as their suggestions in the original paper.

\begin{table*}[hbp]
    \centering
    \caption{The hyper-parameters used in different datasets.}
    \label{tab:8}
    \vskip 0.15in
    \begin{small}
    % \begin{tabular}{*{16}{c}}
    \begin{tabular}{c|ccccccc}
    \toprule
    Hyper-parameters & ECL & Traffic & PeMS04 & Exchange & Weather & ILI & ETT \\
    \midrule
    Learning rate & 1e-3 & 5e-2 & 5e-3 & 5e-4 & 1e-3 & 1e-2 & 5e-3 \\
    Batch size & 16 & 16 & 16 & 8 & 16 & 32 & 32 \\
    \bottomrule
    \end{tabular}
    \end{small}
    \vskip -0.1in
\end{table*}

\section{MTS-Mixers Weights Visualization on Different Datasets}\label{ap:4}

\begin{figure}[h]
\vskip -0.1in
\centering
\subfigure[Visual analysis of MTS-Mixers on ECL.]{\includegraphics[scale=0.34]{figures/ecl_mixing.pdf}} \
\subfigure[Visual analysis of MTS-Mixers on Traffic.]{\includegraphics[scale=0.34]{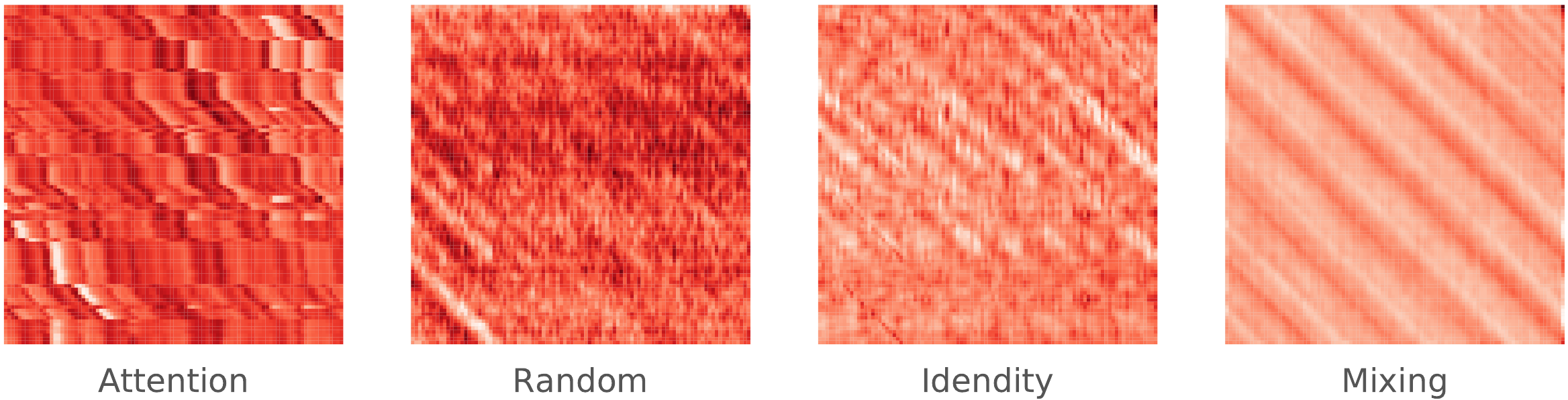}} \
\subfigure[Visual analysis of MTS-Mixers on PeMS04.]{\includegraphics[scale=0.34]{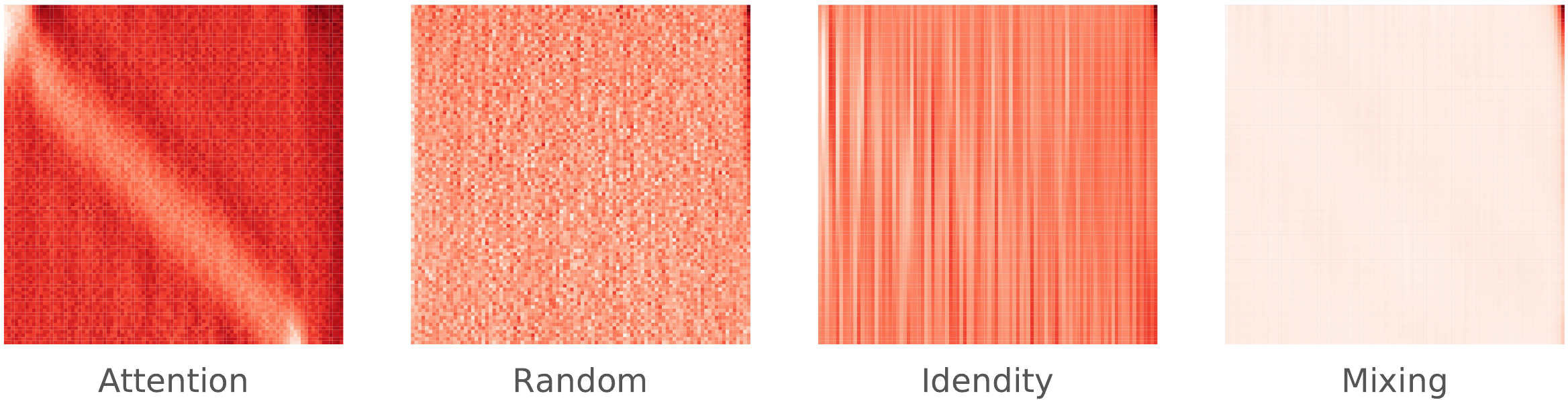}} \
\subfigure[Visual analysis of MTS-Mixers on Exchange.]{\includegraphics[scale=0.34]{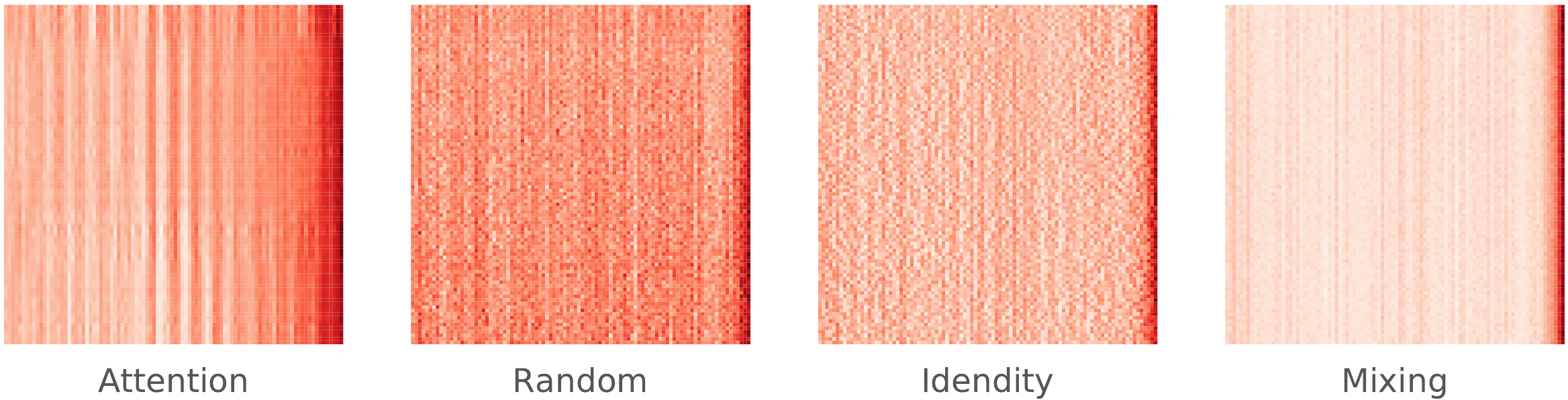}} \
\subfigure[Visual analysis of MTS-Mixers on Weather.]{\includegraphics[scale=0.34]{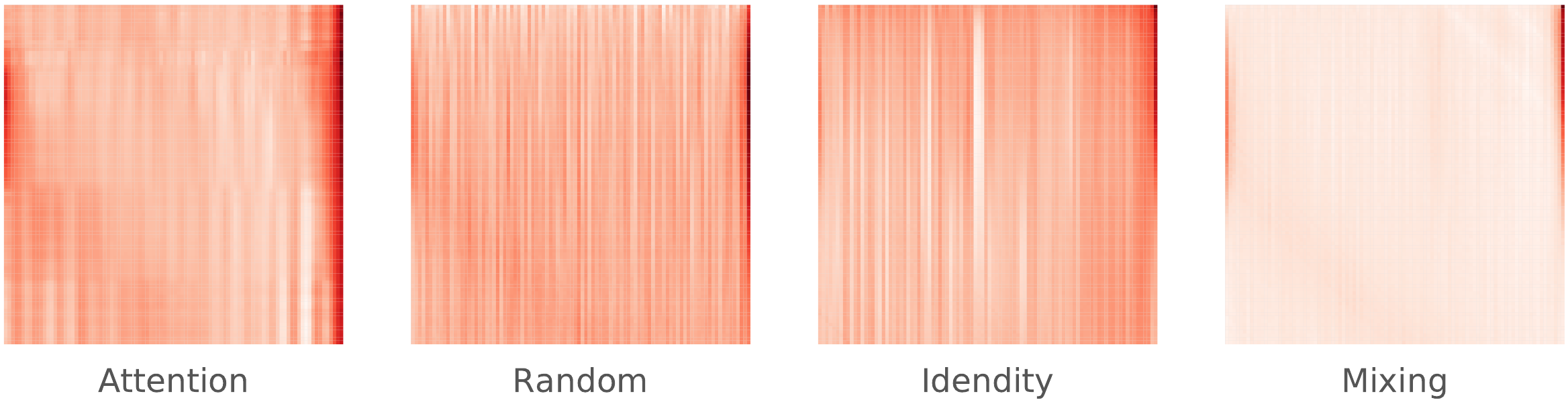}} \
\subfigure[Visual analysis of MTS-Mixers on ILI.]{\includegraphics[scale=0.34]{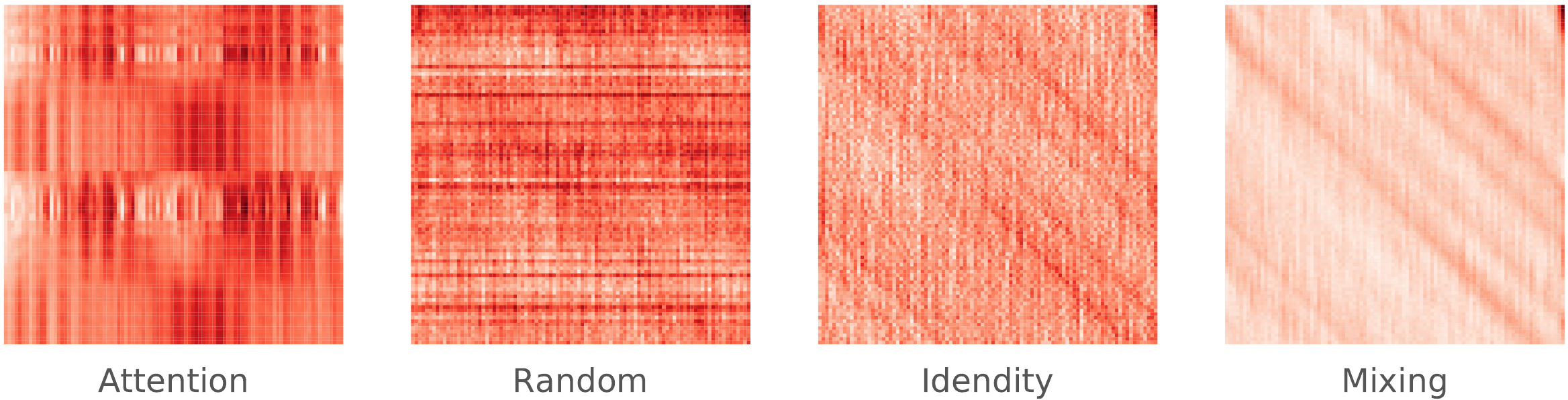}} \
\subfigure[Visual analysis of MTS-Mixers on ETTh1.]{\includegraphics[scale=0.34]{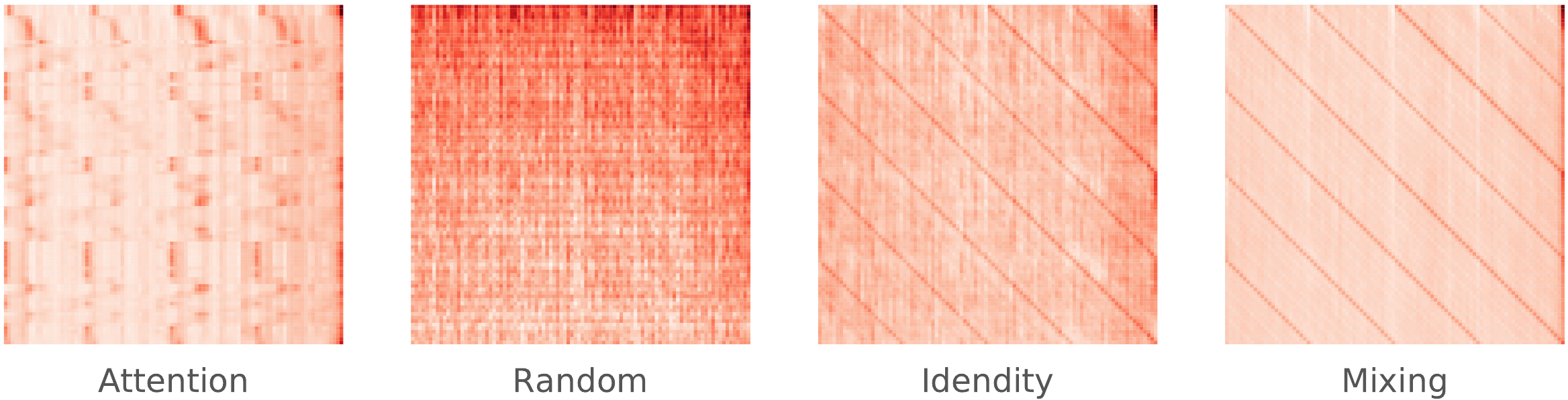}} \
\subfigure[Visual analysis of MTS-Mixers on ETTm1.]{\includegraphics[scale=0.34]{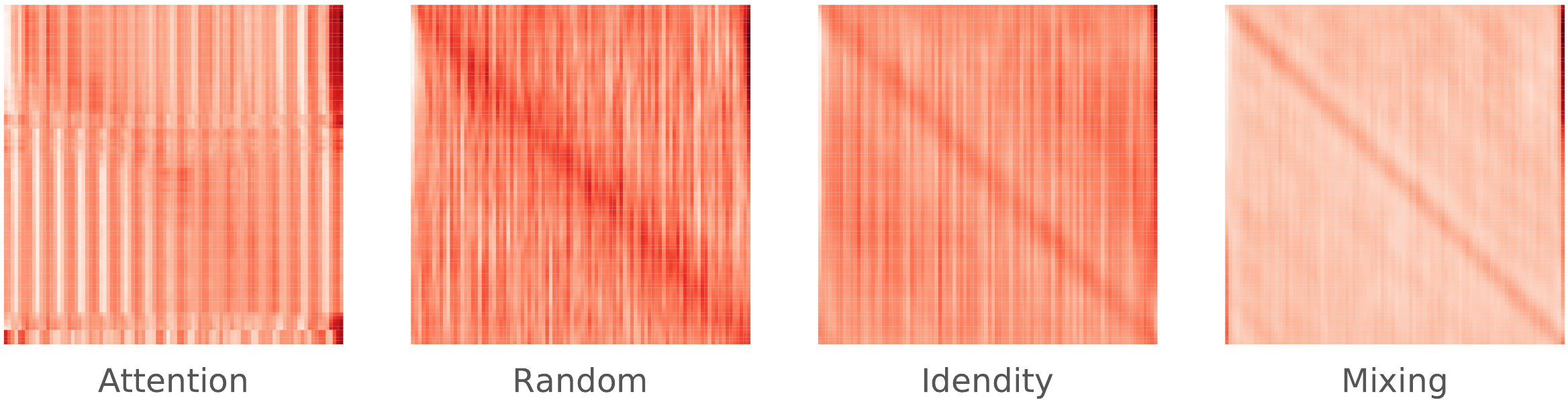}} \
\subfigure[Visual analysis of MTS-Mixers on ETTh2.]{\includegraphics[scale=0.34]{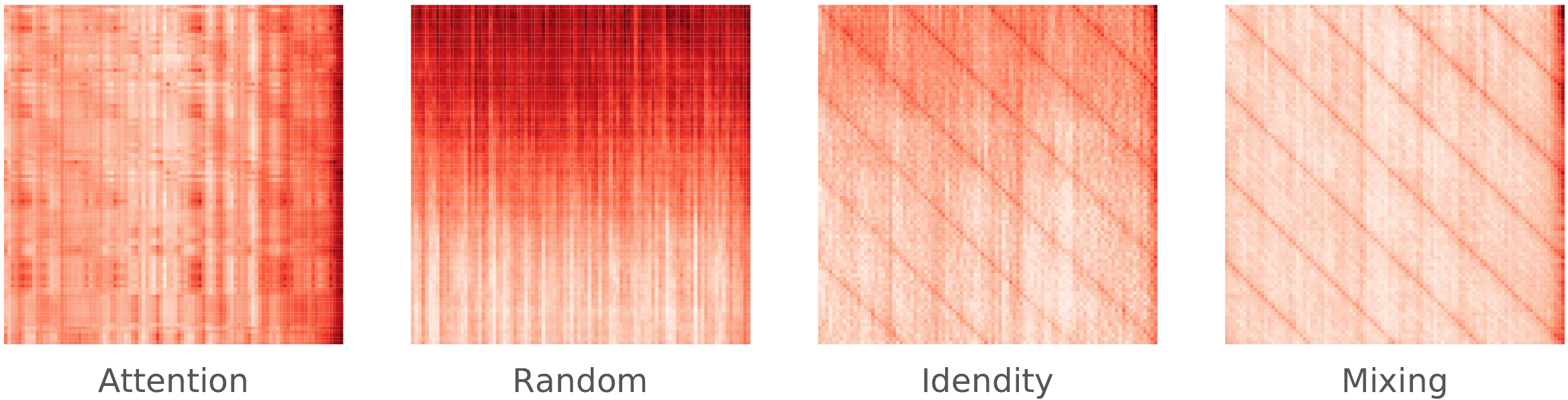}} \
\subfigure[Visual analysis of MTS-Mixers on ETTm2.]{\includegraphics[scale=0.34]{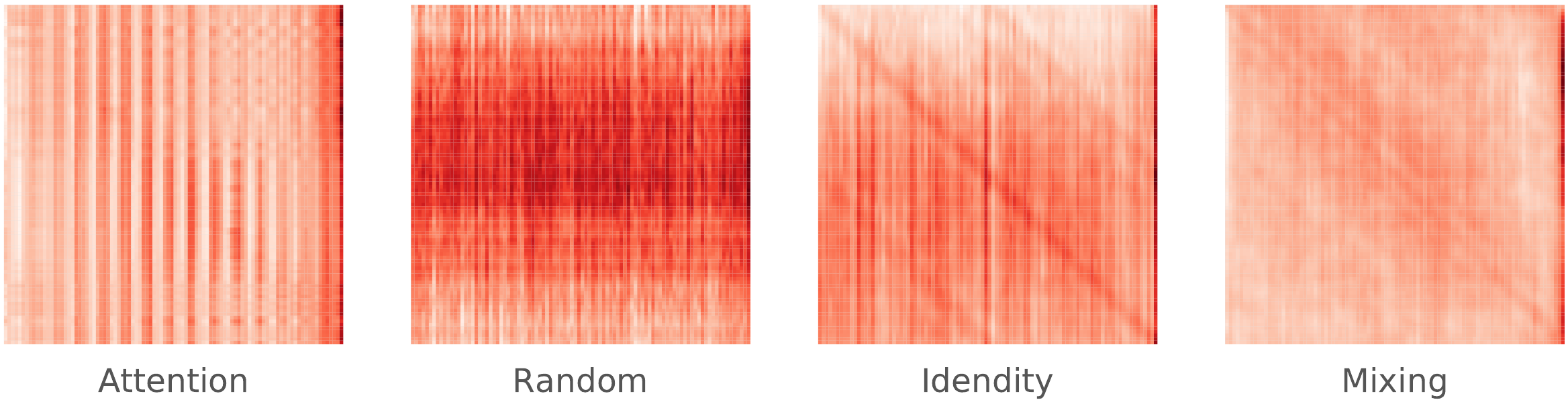}}
\vskip -0.1in
\caption{The visual analysis of MTS-Mixers on different datasets. Only one layer of the stacked block is used for better illustration.}
\vskip -0.1in
\label{fig:8}
\end{figure}

\begin{table*}[htbp]
    \centering
    \caption{From vanilla Transformer to Attention-based MTS-Mixers on the ECL dataset under the 96-96 setting.}
    \label{tab:9}
    \vskip 0.15in
    \begin{small}
    \begin{tabular}{c|cc|c}
    \toprule
    Variants & MSE & MAE & Promotion \\
    \midrule
    Vanilla Transformer & 0.201 & 0.307 & / \\
    Remove the decoder & 0.188 & 0.293 & +6.5\% \\
    Temporal and channel decoupling & 0.179 & 0.284 & +10.9\% \\
    Factorized Attention & 0.165 & 0.272 & +17.9\% \\
    \bottomrule
    \end{tabular}
    \end{small}
    \vskip -0.1in
\end{table*}

\section{From vanilla Transformer to Attention-based MTS-Mixers}\label{ap:5}
Compared with the vanilla Transformer, we remove its decoder and use one linear layer to model the map between learned features and the target sequence. Then we remove the tokenization generally used in Transformer-based frameworks and directly learn temporal information via the self-attention mechanism. Thus, the modeling of temporal and channel interaction is decoupled. Finally, we apply our proposed temporal factorization strategy to the attention, which is an attention-based MTS-Mixer. Table~\ref{tab:9} provides the forecasting results from vanilla Transformer to attention-based MTS-Mixer on the ECL dataset, proving the effectiveness of these modifications.

\section{The Impact of Positional Encoding}
\begin{figure}[ht]
\vskip 0.1in
\centering
\subfigure[MSE of MTS-Mixers w/ or w/o PE.]{\includegraphics[scale=0.38]{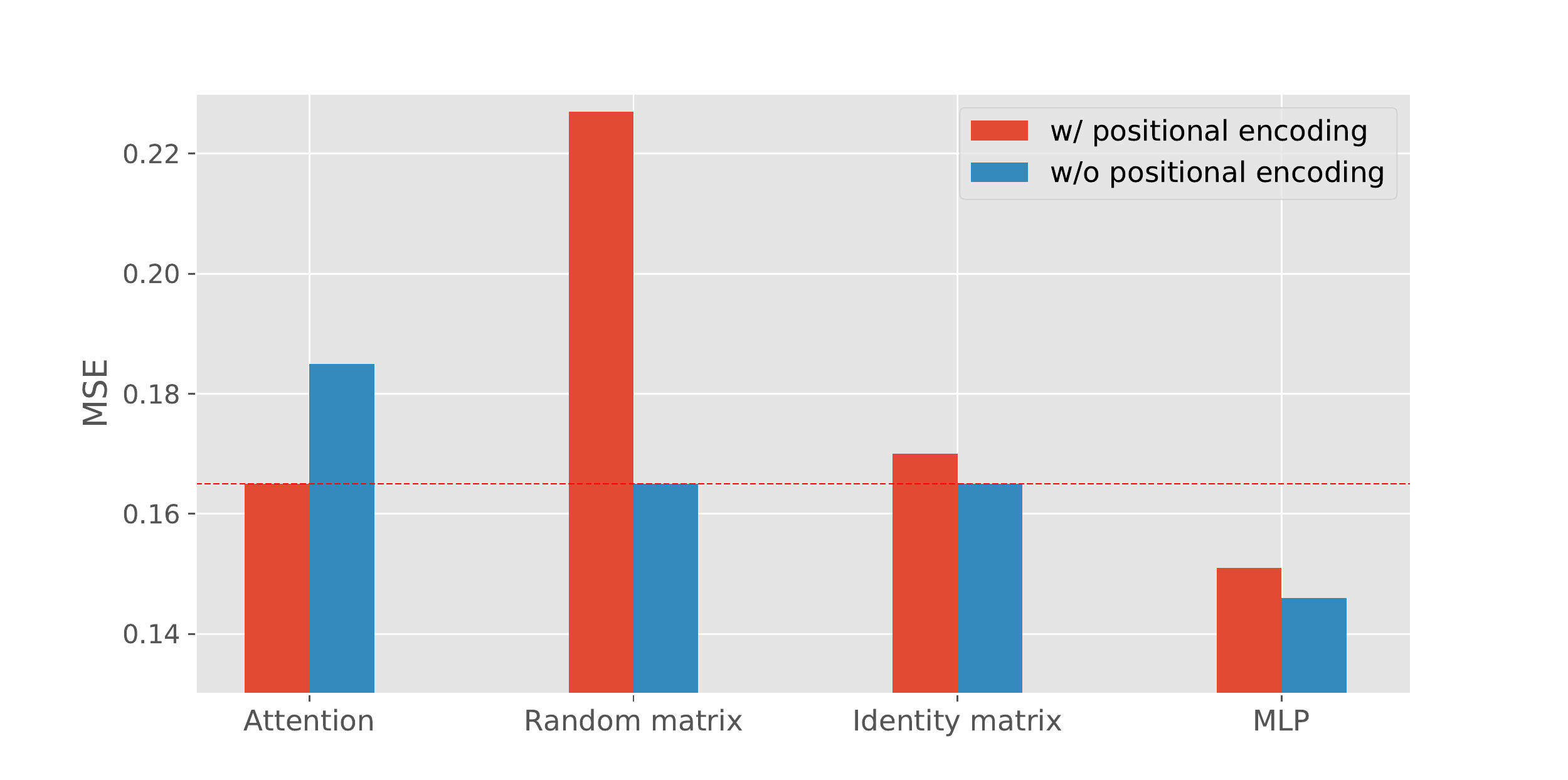}}\quad
\subfigure[MAE of of MTS-Mixers w/ or w/o PE.]{\includegraphics[scale=0.38]{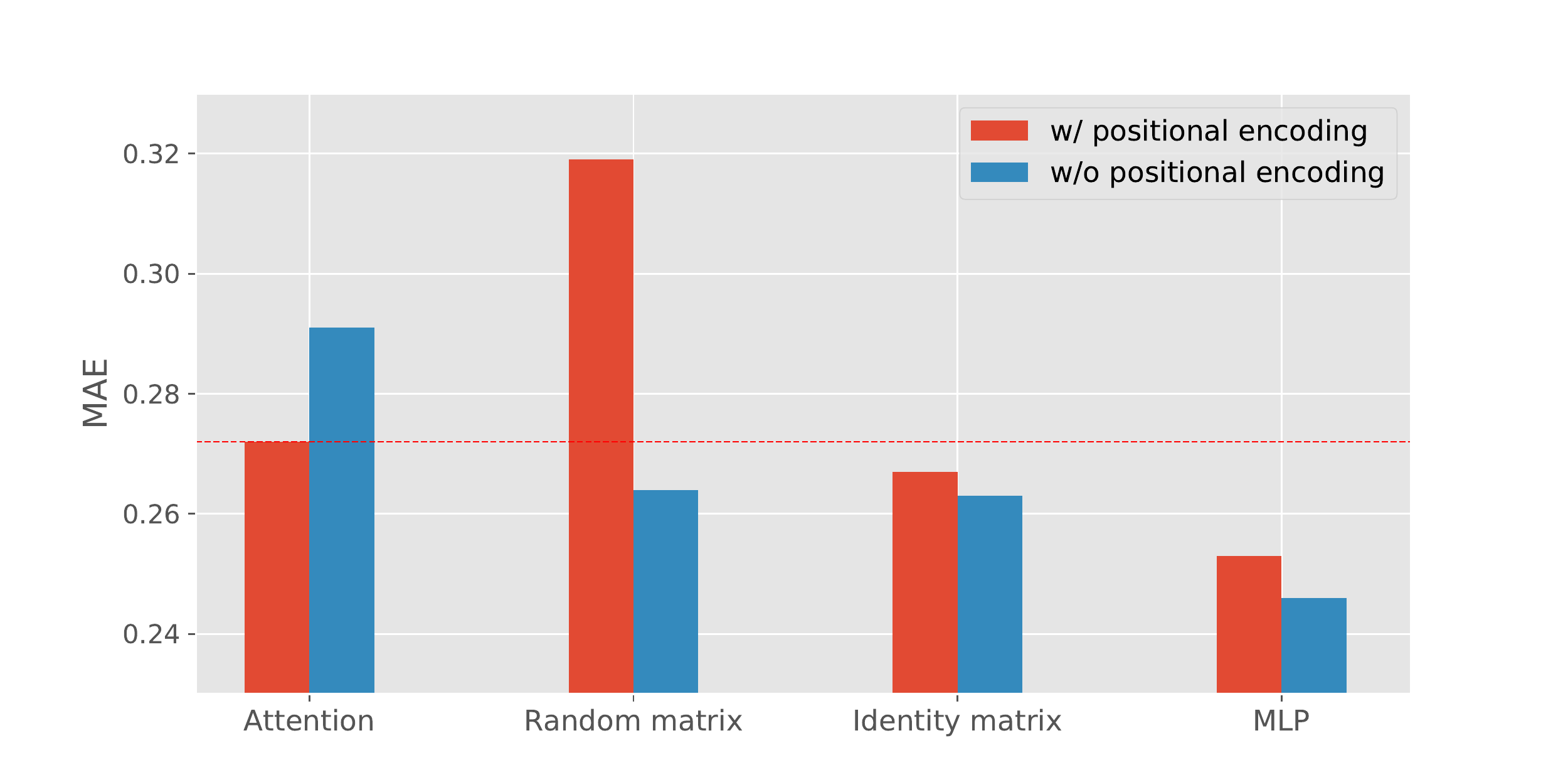}}
\vskip -0.1in
\caption{The results of MTS-Mixers with or without positional encoding on the ECL dataset under the 96-96 setting. The form of positional encoding follows~\cite{Zhou2021InformerBE}. Lower MSE and MAE mean better forecasting performance.}
\label{fig:9}
\end{figure}

Although our proposed attention-based MTS-Mixer outperforms other Transformer-based methods, there is still a gap between attention and matrix-based or MLP-based MTS-Mixer. We argue that it is the distortion caused by positional encoding to capture temporal information. Figure~\ref{fig:9} shows the results of different MTS-Mixers with or without positional encoding on the ECL dataset. Only attention-based MTS-Mixer require positional encoding, while matrix-based and MLP-based MTS-Mixers degrade performance if adding ordering information.

\section{Forecasting Showcases}

\begin{figure}[!hbp]
\vskip 0.1in
\centering
\subfigure[Given 96 to predict 96 steps on the ECL.]{\includegraphics[scale=0.5]{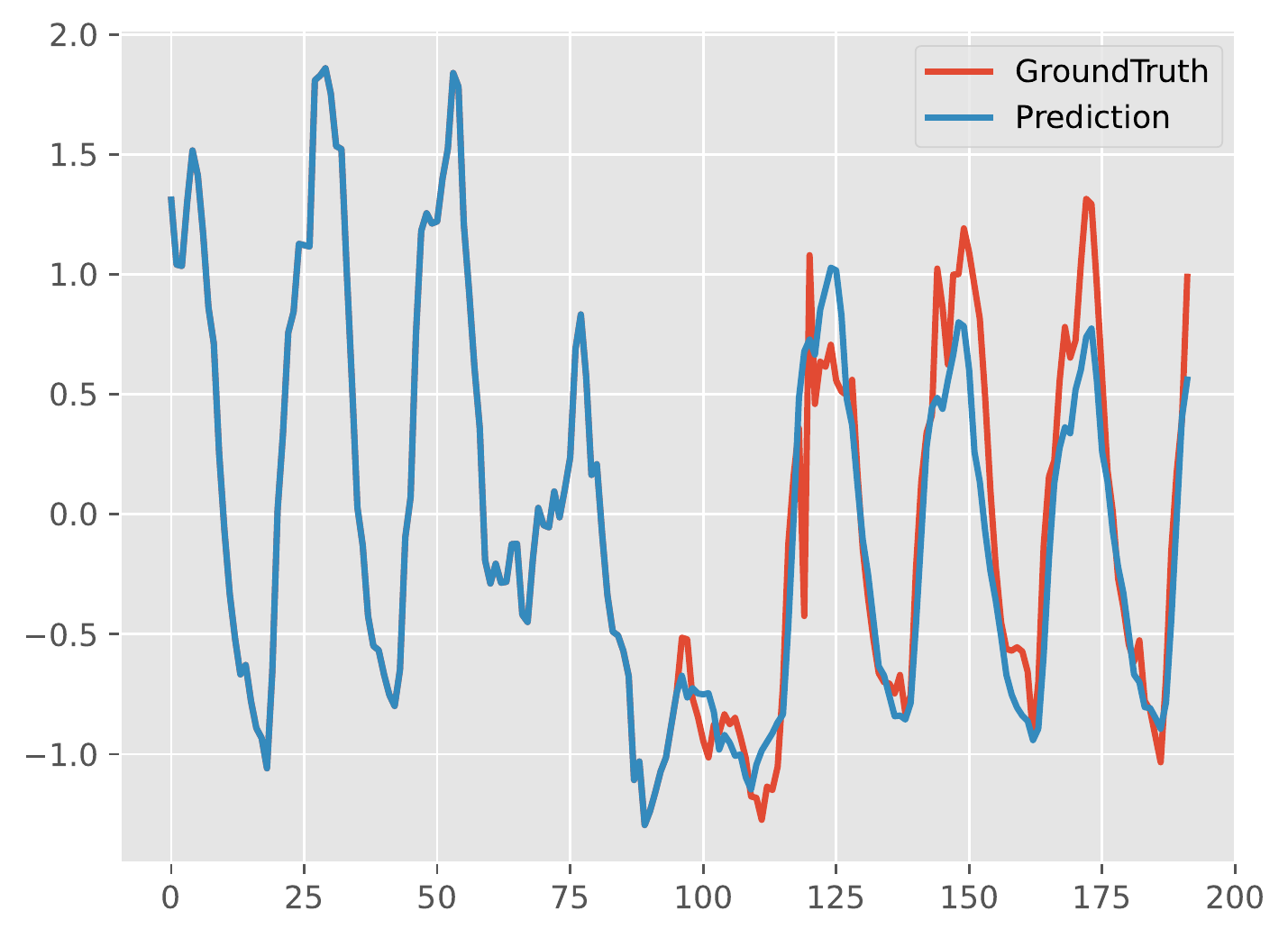}}\qquad\qquad
\subfigure[Given 96 to predict 336 steps on the ECL.]{\includegraphics[scale=0.5]{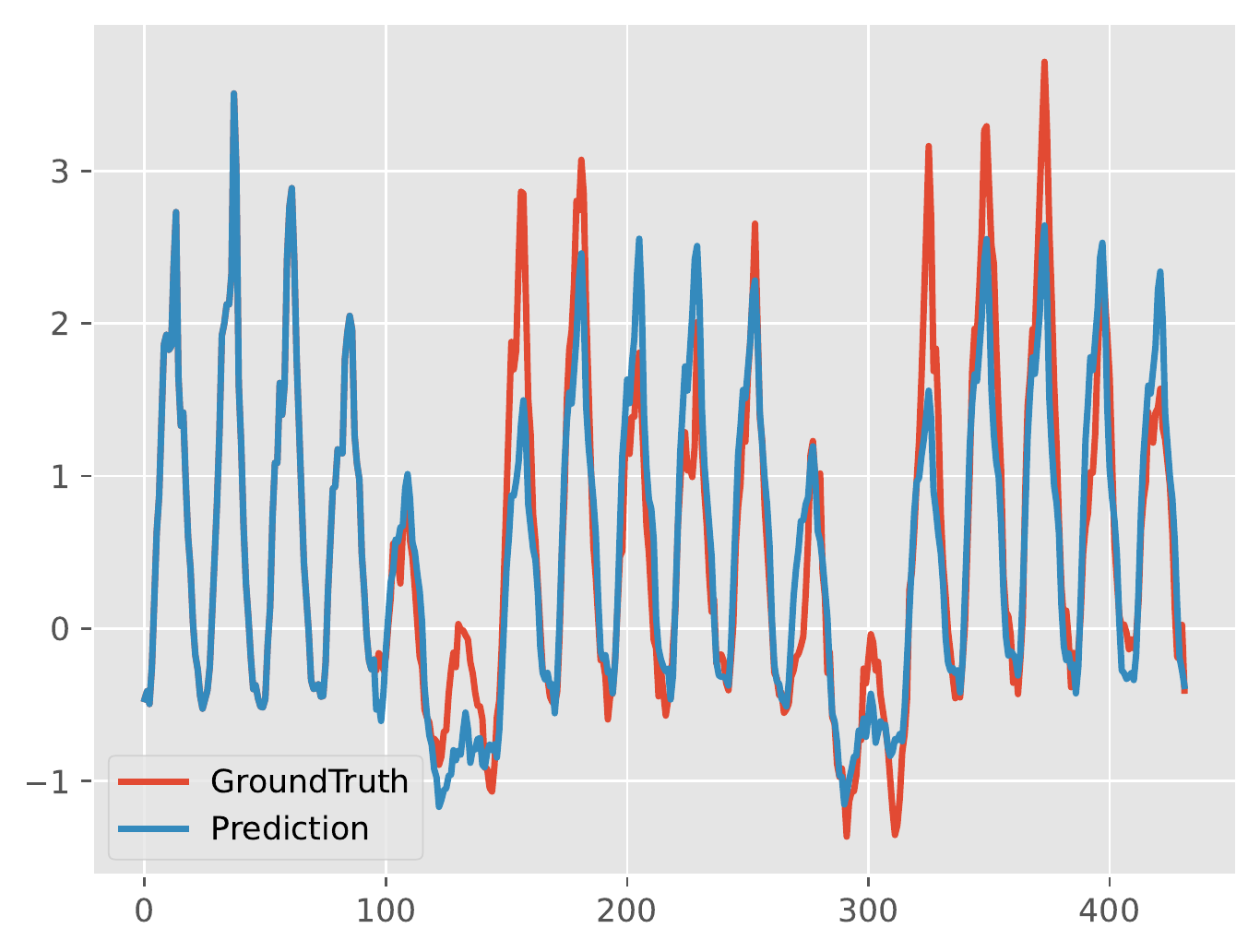}}\
\subfigure[Given 96 to predict 96 steps on the PeMS04.]{\includegraphics[scale=0.5]{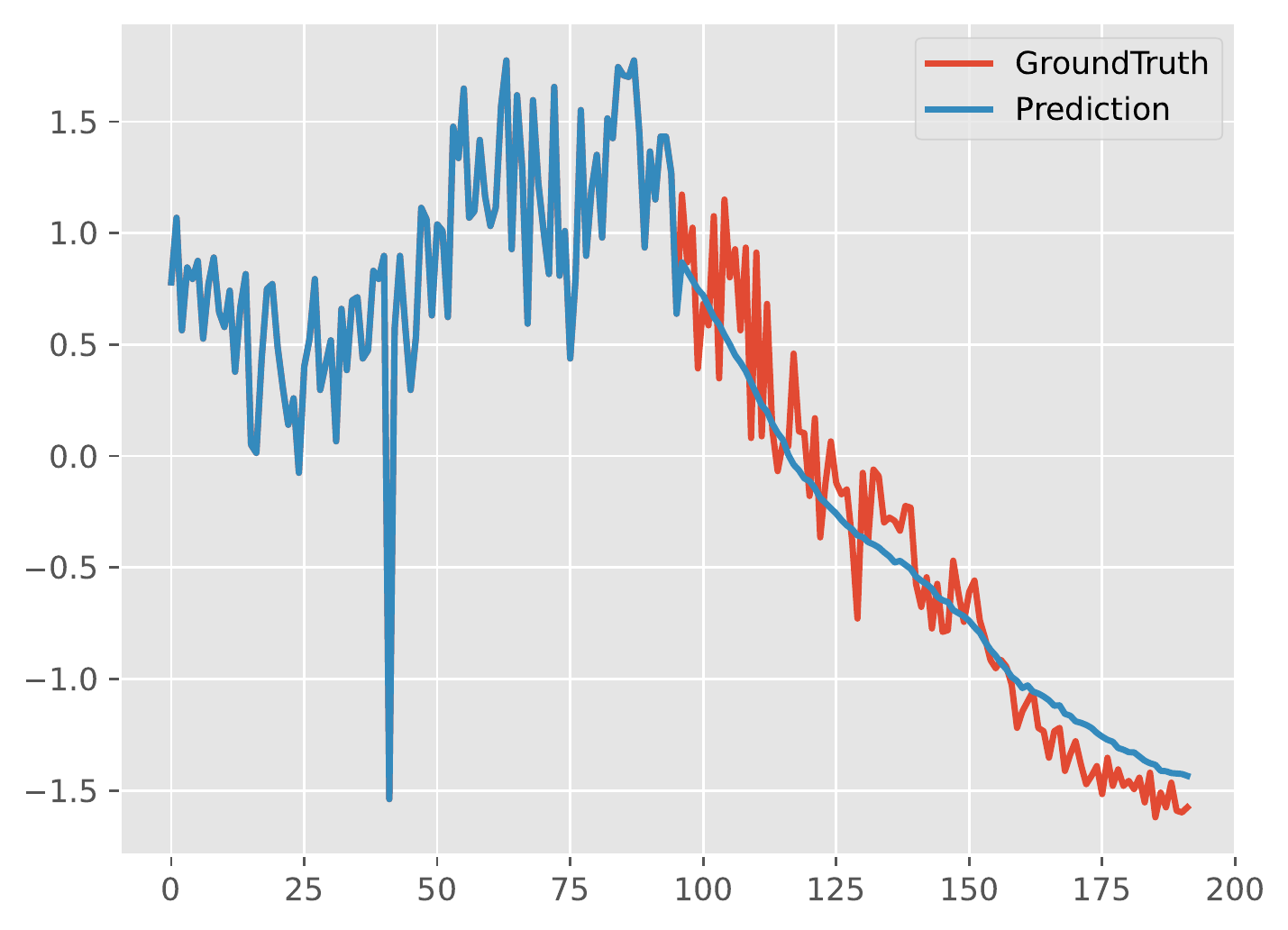}}\qquad\qquad
\subfigure[Given 96 to predict 336 steps on the PeMS04.]{\includegraphics[scale=0.5]{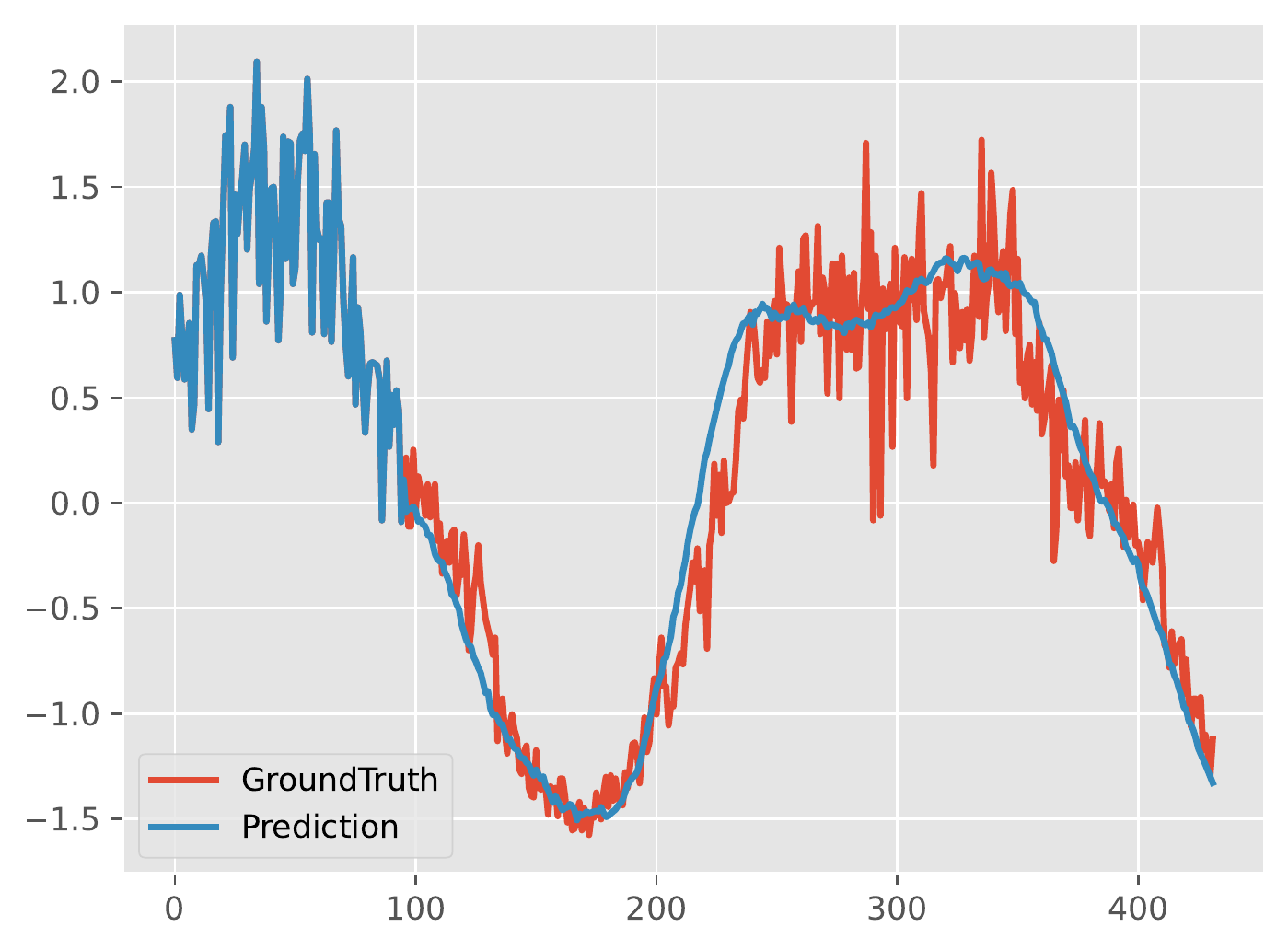}}
\vskip -0.1in
% \caption{Forecasting showcases.}
% \label{fig:10}
\end{figure}

\end{document}